\documentclass[10pt,twocolumn,letterpaper]{article}

\usepackage[pagenumbers]{cvpr} % To force page numbers, e.g. for an arXiv version

\definecolor{cvprblue}{rgb}{0.21,0.49,0.74}
\usepackage[pagebackref,breaklinks,colorlinks,allcolors=cvprblue]{hyperref}

\usepackage{algorithm}
\usepackage{algorithmic}
\usepackage{graphicx}
\usepackage{subcaption}
\usepackage{multirow}
\usepackage{arydshln}
\usepackage[utf8]{inputenc}
\usepackage{amsfonts} 
\usepackage{amsmath} 
\usepackage{booktabs}
\usepackage{pifont}
\usepackage[table]{xcolor}
\definecolor{lightcyan}{RGB}{217, 230, 240} 
\usepackage{arydshln}

\def\paperID{6032} % *** Enter the Paper ID here
\def\confName{CVPR}
\def\confYear{2026}

\title{OWT: A Foundational Organ-Wise Tokenization Framework for Medical Imaging}

\author{Sifan Song$^{1\thanks{Co-first authors.}}$, Siyeop Yoon$^{1\footnotemark[1]}$, Pengfei Jin$^{1}$, Sekeun Kim$^{1}$, Matthew Tivnan$^{1}$, Yujin Oh$^{1}$, Runqi Meng$^{1}$, \\ 
Ling Chen$^{1}$, Zhiliang Lyu$^{1}$, Dufan Wu$^{1}$, Ning Guo$^{1}$, Xiang Li$^{1\footnotemark[2]\thanks{Corresponding authors: Contact {\tt\small li.quanzheng@mgh.harvard. edu} }}$, Quanzheng Li$^{1\footnotemark[2]}$ \\
$^1$ \small Center for Advanced Medical Computing and Analysis (CAMCA), \\ 
     \small Massachusetts General Hospital and Harvard Medical School, Boston, USA\\
}

\begin{document}
\maketitle
\begin{abstract}
Recent advances in representation learning often rely on holistic embeddings that entangle multiple semantic components, limiting interpretability and generalization. These issues are especially critical in medical imaging, where downstream tasks depend on anatomically interpretable features. To address these limitations, we propose an Organ-Wise Tokenization (OWT) framework with a Token Group-based Reconstruction (TGR) training paradigm. Unlike conventional approaches, OWT explicitly disentangles an image into separable token groups, each corresponding to a distinct organ or semantic entity. Our design ensures each token group encapsulates organ-specific information, boosting interpretability, generalization, and efficiency while enabling fine-grained control for targeted clinical applications. Experiments on CT and MRI datasets demonstrate OWT's power: it not only achieves strong performance on standard tasks like image reconstruction and segmentation, but also unlocks novel, high-impact clinical capabilities including organ-specific tumor identification, organ-level retrieval and semantic-level generation, without requiring any additional training. These findings underscore the potential of OWT as a foundational framework for semantically disentangled representation learning, offering broad scalability and a new perspective on how representations can be leveraged.
\end{abstract}    
\section{Introduction}
\label{sec:intro}

\begin{figure}[t]
\centering
\includegraphics[width=0.49\textwidth]{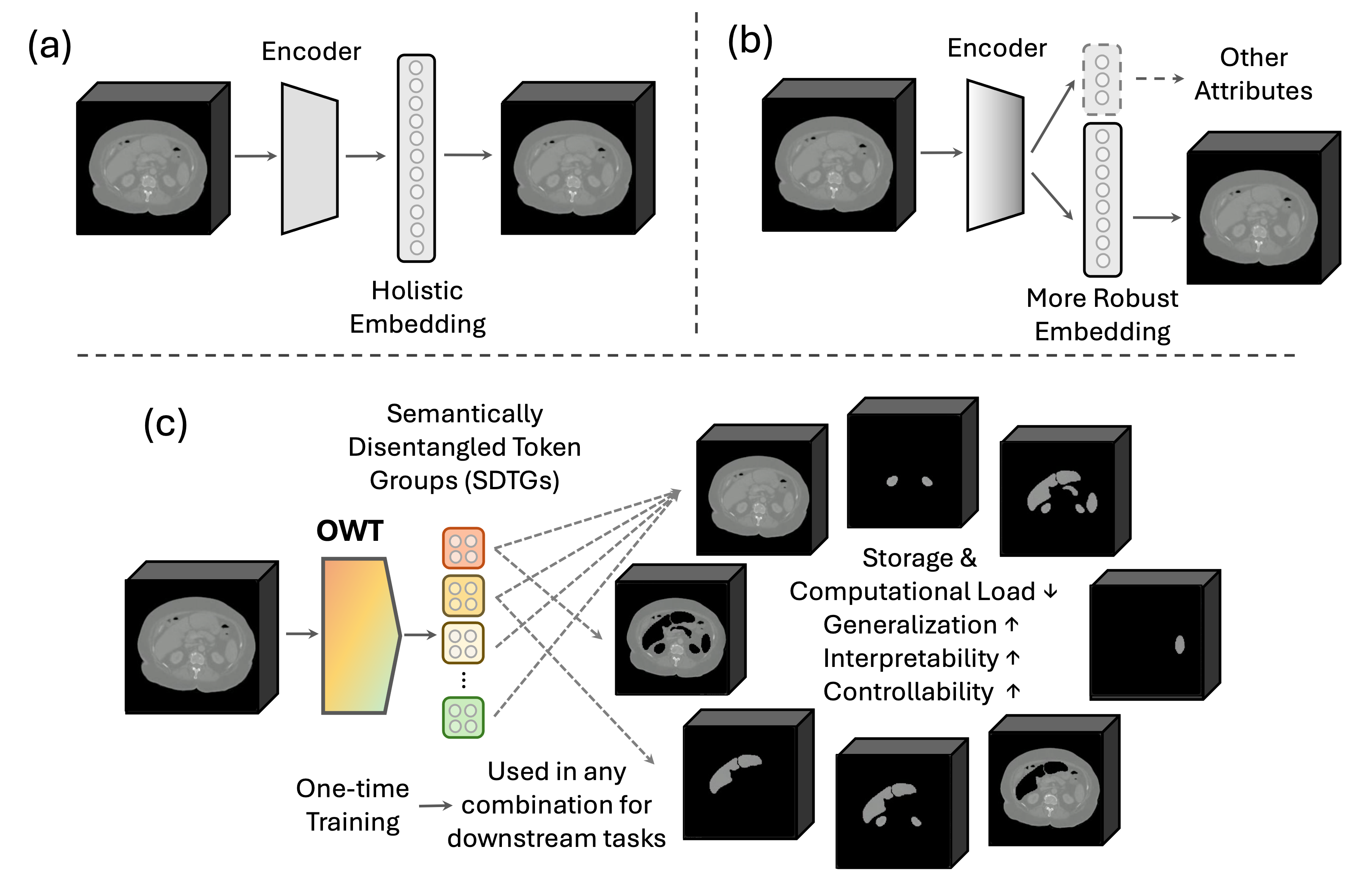}
\caption{Objective comparison of (a) holistic embedding-based representation learning, (b) disentangled representation learning, and (c) our OWT framework. Unlike prior methods, our foundational tokenization framework enables the extraction of disentangled token groups containing semantically meaningful organ-specific representations. These token groups can be leveraged separately or in combination for downstream analyses, such as semantic-level reconstruction, segmentation, and retrieval.} 
\label{fig1}
\end{figure}

Representation learning has advanced rapidly in recent years, with growing impact across both natural and medical imaging domains, enabling more effective feature extraction and improved performance across a wide range of downstream tasks. Building on this progress, a variety of self-supervised and task-guided training frameworks have been widely adopted, typically relying on holistic embeddings that capture aggregated information for an entire image without explicitly separating semantically distinct components.~\cite{voulodimos2018deep, geirhos2020shortcut, wang2024disentangled} (Fig.\ref{fig1}-a). Similar trends appear in the medical imaging domain, where holistic representations have also been extensively explored for tasks like disease diagnosis~\cite{shen2017deep}, organ segmentation~\cite{ronneberger2015u, cao2022swin, hatamizadeh2021swin}, and anatomical structure analysis~\cite{mortada2023segmentation, weston2019automated}. Although holistic embeddings have shown strong performance, they remain globally mixed within a latent space, entangling information across multiple organs and background regions without explicit semantic separation. This mixed representation poses significant challenges for where fine-grained interpretability and a semantically targeted focus are critically important~\cite{schutte2021using, cetin2023attri, wang2024disentangled}.

On the other hand, disentangled representation learning (DRL) aims to decompose data into semantically meaningful factors, each capturing an independent aspect of the underlying variation~\cite{wang2024disentangled} (Fig.\ref{fig1}-b). DRL isolates latent factors (\textit{e.g.}, shape, color, orientation) in natural images~\cite{burgess2018understanding, chen2016infogan, xiao2017dna, zhu2018visual} or modality attributes in medical images~\cite{ouyang2021representation, chartsias2019disentangled}, thereby improving explainability and enabling controllable feature manipulation. However, most existing DRL approaches rely on purely data-driven or perturbation-driven strategies, requiring extensive modeling assumptions or specialized regularizers to ensure each factor corresponds to a semantically coherent concept~\cite{locatello2019challenging, wang2024disentangled}. Consequently, existing DRL approaches suffer from limited alignment with real-world semantics, weak spatial correspondence, and poor scalability (More details on Related Work are provided in \textit{Appendix~\ref{sec:related}}).

While DRL offers a promising direction, both holistic embeddings and existing disentanglement approaches remain limited in medical imaging scenarios, where organ-level interpretability and analysis are important (\textit{e.g.}, anatomical analysis and surgical planning). These limitations manifest in two key aspects: \textbf{(a)} \textit{a lack of interpretability and generalizability}, as multiple semantic cues are often entangled without alignment to real-world concepts, making it difficult to isolate relevant structures from confounding factors (\textit{e.g.}, background); and \textbf{(b)} \textit{limited efficiency and controllability}, as these approaches operate on fused representations rather than isolated organ-level features, restricting the ability to selectively manipulate or analyze distinct semantic factors. These limitations motivate the development of our novel representation learning framework that explicitly tokenizes images at the organ or semantic level. 

In this paper, we propose a foundational \textbf{Organ-Wise Tokenization (OWT)} framework and a novel \textbf{Token Group-based Reconstruction (TGR)} training paradigm (Fig.~\ref{fig1}-c). The OWT framework extends a standard encoder-decoder architecture, such as Masked Autoencoders (MAE) or VQGAN, by incorporating two specialized modules: {\textbf{Organ Collector}} and the {\textbf{Adaptive Holistic Embedding Restorer (AHER)}}, which are inspired by prior work on semantic tokenization~\cite{ryoo2021tokenlearner, esser2021taming, fayyaz2022adaptive}. 

The Organ Collector integrates holistic embeddings into semantically meaningful \textbf{Token Group (TG)}, while AHER enables to reconstruct holistic representations from any number of token groups. This design enables a transition pathway from holistic \(\rightarrow\) semantic \(\rightarrow\) holistic embeddings, facilitating both organ-specific processing and reconstruction. To ensure that the extracted semantic embeddings function as true Semantically Disentangled Token Groups (\textbf{SDTGs}), the proposed TGR training paradigm explicitly enforces correspondence between each token group and its target organ. These components constitute a unified framework that encodes organ-specific semantics into modular and reusable TGs, enabling task-adaptive control and restricting computation to relevant semantic regions without additional training, thereby enhancing interpretability and eliminating redundant processing.

\begin{figure*}
\centering
\includegraphics[width=1.8\columnwidth]{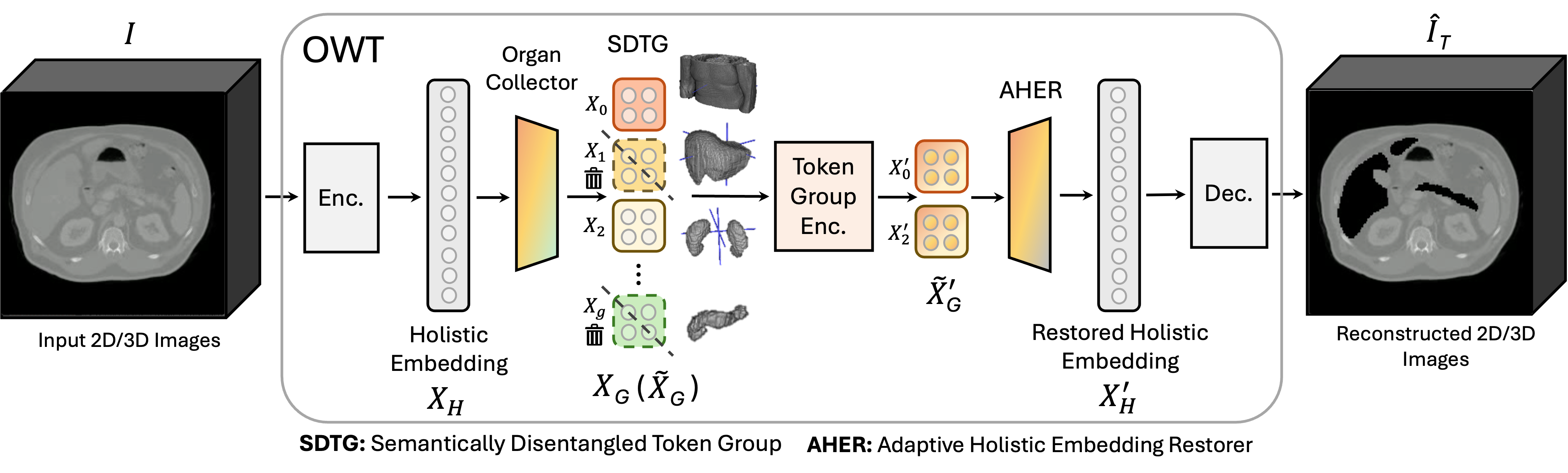}
\caption{Overall architecture of Organ-Wise Tokenization (OWT). The input 2D or 3D images are first encoded into holistic embeddings (\(X_H\)). Then, the holistic tokens are disentangled and embedded in an organ-wise manner using the Organ Collector, forming Semantically Disentangled Token Groups (SDTGs, \(X_G\)). The SDTGs with solid lines represent randomly retained token groups (\(\tilde{X}_G\)). The Token Group Encoder further processes these retained groups, capturing both inter- and intra-organ relationships, and transforming them into \(\tilde{X}'_G\). Finally, an Adaptive Holistic Embedding Restorer and a decoder sequentially restores the holistic representation (\(X'_H\)) and reconstructs the final output (\(\hat{I}_T\)), enabling semantic-level reconstruction.} \label{FigArch}
\end{figure*}

\begin{figure}[]
\centering
\includegraphics[width=0.42\textwidth]{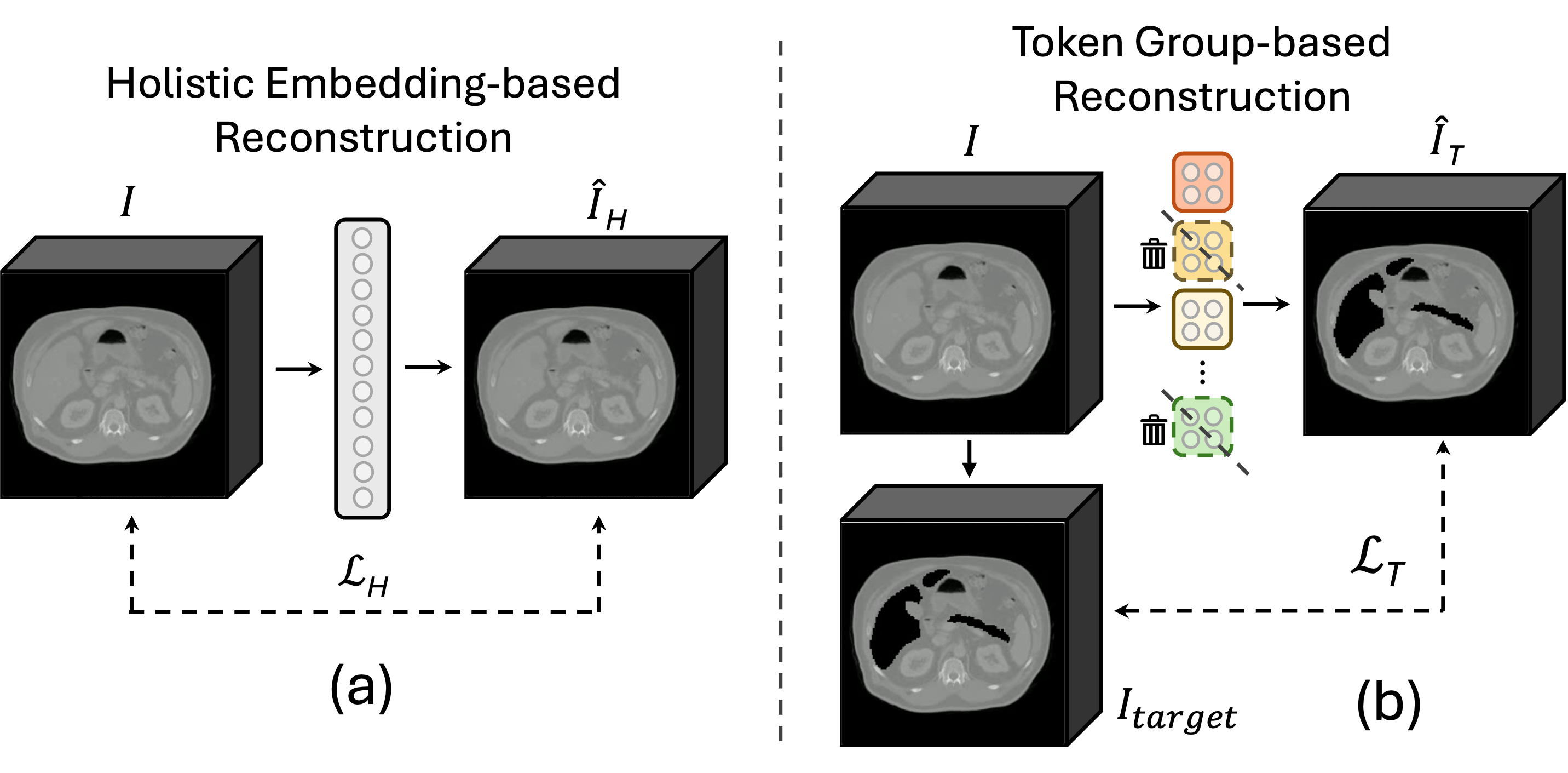}
\caption{Comparison of the overall process of holistic embedding-based and token group-based reconstruction.} 
\label{figTSG}
\end{figure}

We evaluate the organ-wise embeddings on downstream tasks (\eg, image reconstruction and segmentation). Across multiple modalities (CT, two MRI sequences, and a facial dataset), our method consistently outperforms strong baselines, including VAE~\cite{kingma2013auto}, MAE~\cite{he2022masked}, and state-of-the-art task-specific segmentation architectures such as UNet++~\cite{zhou2018unet++}, AttU\_Net~\cite{oktay2018attention}, TransUNet~\cite{chen2021transunet}, Swin-UNet~\cite{cao2022swin}, and CMUNet~\cite{tang2023cmu}. Beyond these general tasks, OWT unlocks capabilities that conventional models cannot provide: organ-level tumor identification and retrieval, semantic-level generation, and flexible combination of organ embeddings to adapt to heterogeneous downstream requirements. This demonstrates that the semantic disentanglement enabled by our OWT framework can be leveraged for broader real-world applications, offering a new perspective on how representations can be constructed and utilized.

In summary, our contributions are listed as follows:
\begin{itemize}
    \item \textbf{We introduce OWT, the first organ-wise foundational framework} that, paired with a novel TGR training paradigm, encapsulates organ-specific semantics into semantically disentangled token groups (SDTGs) within a single training process.
    \item The learned SDTGs are highly condensed yet rich in information for each target organ, enhancing the \textit{interpretability} and \textit{generalizability} of the latent embeddings.
    \item The SDTGs can be selectively combined and reused, yielding improved \textit{efficiency} and fine-grained \textit{controllability}. This unlocks unique organ-level downstream tasks, such as organ-specific retrieval and identification, which are highly demanded in the medical imaging domain.
    \item Our experiments demonstrate significant \textit{scalability}, as OWT adaptively disentangles semantic information across diverse modalities and domains. This includes a real-world facial dataset, highlighting its potential for applications beyond medical imaging.
\end{itemize}
%%%%%%%%%%%%%%%%%%%%%%%%%%%%%%%%%%%%%%%%%%%%%%%%%%%%%%%%%%%%%%%%%%%%%%%%%%%%%%%%%%%%%
%%%%%%%%%%%%%%%%%%%%%%%%%%%%%%%%%%%%%%%%%%%%%%%%%%%%%%%%%%%%%%%%%%%%%%%%%%%%%%%%%%%%%

%%%%%%%%%%%%%%%%%%%%%%%%%%%%%%%%%%%%%%%%%%%%%%%%%%%%%%%%%%%%%%%%%%%%%%%%%%%%%%%%%%%%%
%%%%%%%%%%%%%%%%%%%%%%%%%%%%%%%%%%%%%%%%%%%%%%%%%%%%%%%%%%%%%%%%%%%%%%%%%%%%%%%%%%%%%

\section{Method}
\label{sec:meth}

We propose Organ-Wise Tokenization (OWT) with Token Group-based Reconstruction (TGR): OWT encodes each organ into a semantically disentangled token group (SDTG) that can be used alone or composed for downstream tasks.

Overall, the proposed OWT framework comprises five components (Fig.~\ref{FigArch}): an encoder, Organ Collector, Token Group Encoder, Adaptive Holistic Embedding Restorer (AHER), and a decoder. The encoder maps input images to holistic embeddings, while the Organ Collector and TGR enable the target-organ features to be extracted and stored in specific embeddings (\textit{i.e.}, SDTGs). In TGR, we randomly select a subset of the predefined token groups. The retained token groups are fed into the Token Group Encoder to integrate both inter-organ and intra-organ information. Next, AHER enables any number of retained token groups to be converted back into a holistic embedding. Finally, the decoder reconstructs an image of the same size as the input.

\iffalse
Each component is detailed in the subsections below.
\fi

\subsection{Encoder and Decoder in OWT}
The encoder and decoder in OWT follow standard holistic representation learning architectures, so they can be Transformer-based or (fused) CNN-based, as in 2D/3D MAE~\cite{he2022masked, feichtenhofer2022masked} or 2D/3D VQGAN~\cite{esser2021taming, feichtenhofer2022masked}. This design choice allows the framework to be applied to either 2D or 3D images with minimal modification. For a 3D input volume processed by a Transformer-based encoder, let \(I \in \mathbb{R}^{D \times H \times W \times C}\). The encoder ($\mathbf{Enc}$) produces a holistic embedding $X_H = \mathbf{Enc}(I) \in \mathbb{R}^{(dhw) \times c_e}$, where $D$, $H$, $W$ and $C$ represent the depth, height, width, and channels of the volume, respectively, while $d$, $h$ and $w$ are the corresponding embedded dimensions. The \((dhw)\) represents the number of tokens processed by the Transformer blocks, with each token having an embedding dimension of \(c_e\). This holistic representation is then fed into subsequent modules for more fine-grained organ-wise processing.

%------------------------------------------------------------------------
%------------------------------------------------------------------------
\subsection{Organ Collector}

To effectively extract semantically meaningful features from holistic embedding \(X_H\), the Organ Collector module dynamically assigns importance to different spatial locations or tokens using a token-wise attention matrix \(A_t\).

Let \(X_H \in \mathbb{R}^{(dhw) \times c_e}\) be reshaped to \(\mathbb{R}^{c_e \times d \times h \times w}\) before being fed into the Organ Collector (Fig.~\ref{FigArch}). We then employ a token-wise attention mechanism, defined as:
\begin{equation}
  A_{t} = \text{Softmax} (\text{Flatten}(\alpha(X_H)))
  \;\in\; 
  \mathbb{R}^{N_g \times (hw)},
  \label{equa:collector1}
\end{equation}
\begin{equation}
  X_G = A_{t}\,\text{Flatten}(\gamma(X_H))^{T}
  \;\in\;
  \mathbb{R}^{N_g \times c_e},
  \label{equa:collector2}
\end{equation}
where \(N_g\) denotes the number of semantic-based tokens, and both \(\alpha\) and \(\gamma\) are linear layers. Concretely, \(\alpha\) projects the \((c_e\times d)\)-dimensional input to \(N_g\), then applies \(\text{Flatten}\) and \(\text{Softmax}\) to derive a token-wise attention matrix. This attention matrix is multiplied by \(\gamma(X_H)\) (also flattened and transposed) to obtain \(X_G \in \mathbb{R}^{N_g \times c_e}\). This ensures that each token group captures the most relevant semantic features.

\begin{figure*}
\centering
\includegraphics[width=2.0\columnwidth]{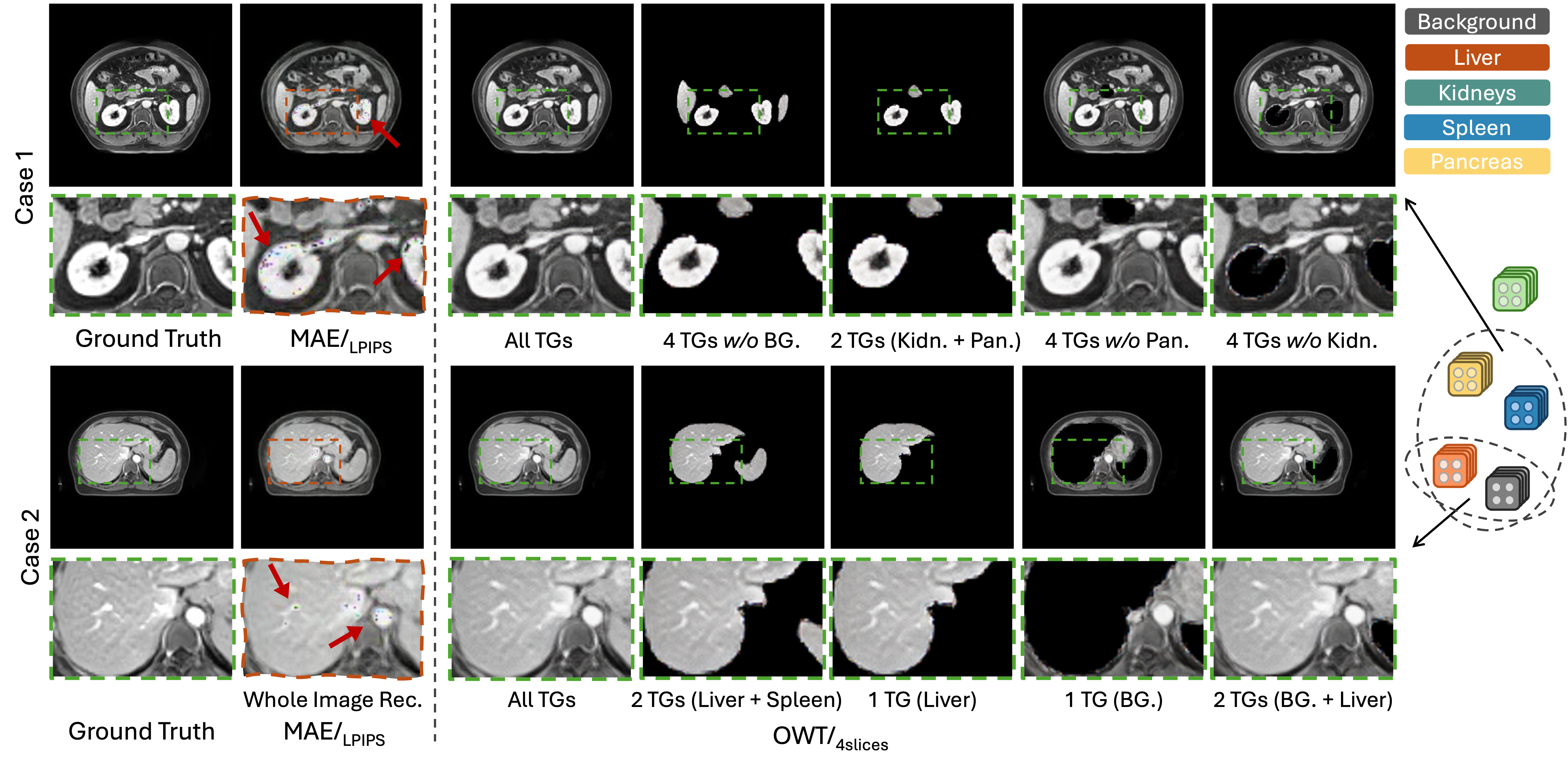}
\caption{Samples of holistic-based and semantic-based medical image reconstruction from an MRI dataset.} 
\label{FigRec1}
\end{figure*}

\subsection{Token Group-based Reconstruction (TGR)}
\label{subsec:TGR}

The TGR training paradigm provides direct supervision of semantic disentanglement to token groups. By enforcing accurate organ-wise feature extraction and reconstruction through the token group-level loss \(\mathcal{L}_T\) (Fig.~\ref{figTSG}-b), TGR preserves both spatial and textural consistency. Consequently, the extracted semantic tokens serve as semantically disentangled token groups (SDTGs). Through TGR, the Organ Collector adaptively extracts organ-specific information into its corresponding SDTG, assigning each token group to a particular organ.

\noindent
\textbf{Definition of Token Group.} Given a dataset \(D\), we have its ground-truth annotations for \(g\) semantic group (such as organs in medical image). The remaining regions are treated as background and assigned to a single additional token group. Thus, we define a total of \((g+1)\) token groups. Let each group contain \(f\) tokens, making the total number of tokens \(N_g = (g+1)\times f\). For example, in Fig.~\ref{FigArch}, \(X_0\), \(X_1\), and \(X_2\) represent the token groups that store liver, kidney, and background information, respectively.

\noindent
\textbf{Random Selection Procedure.} To establish and preserve the correspondence between token groups and their target organs during training, we apply a two-stage random selection.
\textit{\textbf{First}}, {randomly determine the number of retained groups:} Let \(\Omega \sim \mathcal{U}(0,1)\) be a uniform random variable on \([0,1]\). With \((g+1)\) total Token Groups, we define \(\tilde{g} = \bigl\lfloor (g+1) \cdot \Omega \bigr\rfloor\), resulting in an integer \(\tilde{g} \in \{0,1,\dots,g\}\).  
\textit{\textbf{Second}}, {randomly select \(\tilde{g}\) groups:} From the full set of \((g+1)\) token groups, we then randomly pick exactly \(\tilde{g}\) groups to retain and discard the rest. We set each group to contain \(f\) tokens so that each separable and independent token group is in \(\mathbb{R}^{{f} \times c_e}\). We concatenate the \(\tilde{g}\) selected groups along the token dimension, producing the final retained tokens \(\tilde{X}_G \in \mathbb{R}^{{\tilde{N}_g} \times c_e}\) (Fig~\ref{FigArch}), where the final retained token count is \(\tilde{N}_g = \tilde{g} \times f\).

During TGR, for each input image, we remain only areas according to randomly retain token groups based on organ annotations (Fig.~\ref{figTSG}-b), treating the processed image as the target \(I_{target}\) for the token group-based reconstruction loss \(\mathcal{L}_{T}\). This explicit supervision encourages a precise organ-level focus, preserving each organ’s unique spatial and textural details in the retained token groups (\textit{i.e.}, SDTGs). Meanwhile, any randomly excluded token groups are discarded at this stage, ensuring that only the retained groups contribute to reconstruction and thereby enhancing semantic flexibility and computational efficiency. As illustrated in Fig.~\ref{figTSG}, TGR departs from conventional holistic embedding-based reconstruction methods that rely on the latent space for the entire image~\cite{bank2023autoencoders, he2022masked, feichtenhofer2022masked, esser2021taming} (Fig.~\ref{figTSG}-a). By selectively preserving certain token groups, our approach avoids fusing diverse organ features and does not require masked tokens, resulting in improved interpretability, efficiency, and controllability during both reconstruction and subsequent tasks.

After that, we employ a Token Group Encoder ($\mathbf{Enc}_{G}$) composed of six multi-head linear self-attention blocks~\cite{katharopoulos2020transformers}. Let $\tilde{X}'_G = \mathbf{Enc}_{G}(\tilde{X}_G) \in \mathbb{R}^{\tilde{N}_g \times c_e}$ represent the further contextual modeling that captures inter- and intra-organ relationships of the retained token groups.

\subsection{Adaptive Holistic Embedding Restorer}

Instead of relying on conventional masked tokens, our method discards unselected token groups in the random selection procedure, simplifying training and avoiding unnecessary overhead. After the last module, each input can produce a different number of retained token groups (\(\tilde{X}'_G\) in Fig.~\ref{FigArch}), resulting in a variable token count \(\tilde{N}_g\). To convert these tokens back into a representation aligned with the original input size (\textit{i.e.}, the shape of the pre-Organ Collector embedding, \(X_H\)), we propose an Adaptive Holistic Embedding Restorer (AHER). This module dynamically accepts the variable token count \(\tilde{N}_g\) and restores them to a consistent dimension suitable for subsequent reconstruction.

The AHER module is also implemented using a token-wise attention mechanism, defined as:
\begin{equation}
    A'_t = \text{Softmax}(\phi(\tilde{X}'_G)^T) \;\in\; \mathbb{R}^{(dhw) \times \tilde{N_g}} \;,
    \label{equa:DDCL1}
\end{equation}
\begin{equation}
    X'_H = A'_t\psi(\tilde{X}'_G) \;\in\; \mathbb{R}^{(dhw) \times c_e} \;,
    \label{equa:DDCL2}
\end{equation}
where \(\phi(\cdot)\) and \(\psi(\cdot)\) are two linear transformations acting on \(\tilde{X}'_G\), and \((dhw)\) is the flattened spatial dimension we seek to restore (\textit{i.e.}, the depth $\times$ height $\times$ width of the input volume). Specifically, a token-wise attention matrix \(A'_t\) is computed via softmax on the transposed \(\phi(\tilde{X}'_G)\). Next, the features \(\psi(\tilde{X}'_G)\) are re-aligned according to \(A'_t\), yielding a restored holistic embedding \(X'_H\) (Fig.~\ref{FigArch})

Although \(\tilde{N}_g\) varies per input, this operation preserves organ-specific features while ensuring \(X'_H\) remains a fixed \((dhw)\times c_e\) shape, thereby maintaining consistency with the decoder. Finally, we pass \(X'_H\) through a standard decoder~\cite{he2022masked} ($\mathbf{Dec}$) and unpatchify the output, reconstructing the image $\hat{I}_T = \mathbf{Dec}(X'_H) \in \mathbb{R}^{D \times H \times W \times C}$.

As mentioned in Sec.~\ref{subsec:TGR}, our objective is to reduce the discrepancy between the target image \(I_{target}\) and the reconstructed image \(\hat{I}_T\). We combine an L2 term with the LPIPS perceptual loss~\cite{zhang2018unreasonable}
\begin{equation}  
\mathcal{L}_T = \| I_{target} - \hat{I}_T \|_2^2\ + \mathcal{L}_{LPIPS}(I_{target}, \hat{I}_T) \;
  \label{equa:loss}
\end{equation}
to preserves low-level fidelity while capturing high-level perceptual similarity, thereby guiding the network to generate more realistic texture and coherent organ regions.

%%%%%%%%%%%%%%%%%%%%%%%%%%%%%%%%%%%%%%%%%%%%%%%%%%%%%%%%%%%%%%%%%%%%%%%%%%%%%%%%%%%%%
%%%%%%%%%%%%%%%%%%%%%%%%%%%%%%%%%%%%%%%%%%%%%%%%%%%%%%%%%%%%%%%%%%%%%%%%%%%%%%%%%%%%%

\begin{table}[t]  
\centering
\resizebox{1.0\columnwidth}{!}{
\begin{tabular}{lcccccc}
\toprule
 %Tasks & Pre-defined Rec. & Any Comb. Rec. & Segmentation & Any Comb. Retrieval \\
\multirow{2}{*}{Tasks} & Organ-specific & Organ-level & Pre-defined & Any Comb. & {Segment-} \\
& Identification & Retrieval & Rec. & Rec. & ation \\
\midrule
Rec. Nets  & \ding{55} & \ding{55} & \checkmark & \ding{55} & \ding{55} \\
Seg. Nets  & \ding{55} & \ding{55} & \ding{55} & \ding{55} & \checkmark \\
\cellcolor{lightcyan} \textbf{OWT}  & \checkmark & \checkmark & \checkmark & \checkmark & \checkmark \\
\bottomrule
\end{tabular}
}
\caption{Downstream task capabilities after one-time training.}
\label{tab: downcheck}
\end{table}

\section{Datasets and Experimental Setup}
\label{sec:dataandexp}

We conduct experiments on four 3D medical datasets: two CT sets (Abdomen1k~\cite{ma2021abdomenct} and the CT volumes from AutoPet~\cite{Gatidis2022}) and two synthetic MRI sequences (delay and pre-artery) from RAOS dataset~\cite{luo2024rethinking}. These multimodal and multi-sequence datasets validate the effectiveness and robustness of the OWT framework in medical imaging. To further validate that OWT can generalize to different semantic units beyond medical scenarios, we employ the CelebAMaskHQ dataset~\cite{lee2020maskgan}. More details are in \textit{Appendix~\ref{subsec: pre}}, including image preprocessing and data splitting.

All experiments are conducted on six Nvidia H100 GPUs. For fairness, we also add LPIPS loss~\cite{zhang2018unreasonable} to VAE and MAE to improve the quality of reconstruction. Further details of the architecture descriptions and  hyperparameters are in \textit{Appendix~\ref{subsec: defarch}} and \textit{\ref{subsec: exset}}.

%-----------------------------------------------------------------------------------------------------------------------------------
%-----------------------------------------------------------------------------------------------------------------------------------
\begin{table}[t]  
\centering
\resizebox{0.9\columnwidth}{!}{
\begin{tabular}{lcccccc}
\toprule
Methods & Enc. & OC. & TG Enc. & AHER & Dec. & Total \\
\midrule
MAE & 16.66 & - & - & - & 11.11 & 27.77 \\
\cellcolor{lightcyan} \textbf{OWT} & 8.33 & 0.25 & 4.25 & 0.074 & 11.11 & \textbf{24.01} \\
\bottomrule
\end{tabular}
}
\caption{Comparison of computation costs (GFLOPs).}
\label{tab: calculate}
\end{table}

\begin{table}[t]
\centering
\resizebox{0.9\columnwidth}{!}{
\begin{tabular}{lcccc}
\toprule
\multirow{2}{*}{Methods} & Disentangled & {Used} & \multicolumn{2}{c}{Tumor Identification} \\
\cmidrule(lr){4-5}
& Organ Info. & Token Num.& Liver & Kidneys \\
\midrule
MAE & \ding{55} & 196 & 73.5 & 74.7 \\
\cellcolor{lightcyan} \textbf{OWT} & \checkmark & \textbf{20} & \textbf{79.7} & \textbf{81.8} \\
\bottomrule
\end{tabular}
}
\caption{Performance for the tumor identification tasks.}
\label{tab: tumor}
\end{table}

\section{Results}
\label{sec:res}

In this section, we design extensive experiments to explore OWT's unique clinical impacts (Sec.~\ref{subsec:clinical}), and to evaluate performance on several downstream tasks (Sec.~\ref{subsec:down}). Next, we dive into SDTGs and investigate their semantic characteristics (Sec.~\ref{subsec:look}). We also explore the potential of our OWT to address the common issue of limited medical annotations (Sec.~\ref{sec: percelabel}). Finally, we demonstrate that the OWT framework can also be employed in natural image applications (Sec.\ref{subsec:face}), and provide ablation studies in \textit{Appendix~\ref{sec: abl}} to discuss several important hyperparameters of OWT, including encoder/decoder alterations, different token number per organ, and data percentage used for training.

\begin{table*}[t!]
\begin{minipage}[b]{.50\textwidth}
  \centering
\resizebox{1.0\columnwidth}{!}{
\begin{tabular}{lcccccc}
\toprule
\multirow{2}{*}{Abdomen1k} & \multicolumn{3}{c}{Holistic Rec.} & \multicolumn{3}{c}{1 Token Group Rec.} \\
\cmidrule(lr){2-4} \cmidrule(lr){5-7}
 & L2$\downarrow$ & LPIPS$\downarrow$  & SSIM$\uparrow$ & L2$\downarrow$ & LPIPS$\downarrow$ & SSIM$\uparrow$ \\
\midrule
VAE/$_\text{LPIPS}$ & 6.08e-03 & 0.1717 & - & \ding{55}    & \ding{55}   & \ding{55} \\
MAE      &    6.74e-04        &   0.0684      &   0.9688       & \ding{55}    & \ding{55}   & \ding{55}        \\
MAE/$_\text{LPIPS}$ & 6.19e-04   & 0.0403  & 0.9722   & \ding{55}    & \ding{55}   & \ding{55}  \\
\cellcolor{lightcyan} OWT    & 4.16e-04   & 0.0390  & 0.9734   & 3.31e-04   & 0.0073   & 0.9795    \\
\cellcolor{lightcyan} OWT/$_{\text{4-slice}}$ & \textbf{3.69e-04}  & \textbf{0.0371}  & \textbf{0.9755}   & \textbf{2.68e-04}  & \textbf{0.0069}   & \textbf{0.9838}    \\
\toprule
\toprule
\multirow{2}{*}{AutoPet} & \multicolumn{3}{c}{Holistic Rec.} & \multicolumn{3}{c}{1 Token Group Rec.} \\
\cmidrule(lr){2-4} \cmidrule(lr){5-7}
 & L2$\downarrow$ & LPIPS$\downarrow$  & SSIM$\uparrow$ & L2$\downarrow$ & LPIPS$\downarrow$ & SSIM$\uparrow$ \\
\midrule
VAE/$_\text{LPIPS}$ & 3.43e-03 & 0.1204 & - & \ding{55}    & \ding{55}   & \ding{55} \\
MAE      &     2.86e-04       &    0.0452     &     0.9799     & \ding{55}    & \ding{55}   & \ding{55}\\
MAE/$_\text{LPIPS}$ & 2.77e-04   & 0.0255  & 0.9843   & \ding{55}    & \ding{55}   & \ding{55} \\
\cellcolor{lightcyan} OWT    & 1.49e-04   & 0.0232  & 0.9881   & 3.32e-04   & 0.0071   & 0.9778    \\
\cellcolor{lightcyan} OWT/$_{\text{4-slice}}$    & \textbf{1.13e-04}   & \textbf{0.0179}  & \textbf{0.9908}   & \textbf{2.68e-04}   & \textbf{0.0063}   & \textbf{0.9827}    \\
\bottomrule
\end{tabular}
}
\captionof{table}{Performance of image reconstruction on CT datasets.}
\label{tab: recon_ct}
\end{minipage}\quad
\begin{minipage}[b]{.50\textwidth}
  \centering
\resizebox{1.0\columnwidth}{!}{
\begin{tabular}{lcccccc}
\toprule
\multirow{2}{*}{Delay} & \multicolumn{3}{c}{Holistic Rec.} & \multicolumn{3}{c}{1 Token Group Rec.} \\
\cmidrule(lr){2-4} \cmidrule(lr){5-7}
 & L2$\downarrow$ & LPIPS$\downarrow$  & SSIM$\uparrow$ & L2$\downarrow$ & LPIPS$\downarrow$ & SSIM$\uparrow$ \\
\midrule
VAE/$_\text{LPIPS}$ & 9.88e-03 & 0.1070 & - & \ding{55}    & \ding{55}   & \ding{55} \\
MAE      &      1.37e-03      &    0.0465     &     0.9686     & \ding{55}    & \ding{55}   & \ding{55} \\
MAE/$_\text{LPIPS}$ & 1.36e-03   & 0.0389  & 0.9690   & \ding{55}    & \ding{55}   & \ding{55} \\
\cellcolor{lightcyan} OWT    & 7.65e-04   & 0.0256  & 0.9851   & 5.96e-04   & 0.0088   & 0.9762    \\
\cellcolor{lightcyan} OWT/$_{\text{4-slice}}$    & \textbf{5.81e-04}   & \textbf{0.0214}  & \textbf{0.9885}   & \textbf{5.04e-04}   & \textbf{0.0081}   & \textbf{0.9801}    \\
\toprule
\toprule
\multirow{2}{*}{PreArtery} & \multicolumn{3}{c}{Holistic Rec.} & \multicolumn{3}{c}{1 Token Group Rec.} \\
\cmidrule(lr){2-4} \cmidrule(lr){5-7}
 & L2$\downarrow$ & LPIPS$\downarrow$  & SSIM$\uparrow$ & L2$\downarrow$ & LPIPS$\downarrow$ & SSIM$\uparrow$ \\
\midrule
VAE/$_\text{LPIPS}$ & 1.16e-02 & 0.1059 & - & \ding{55}    & \ding{55}   & \ding{55} \\
MAE      &  1.67e-03     &    0.0438     &    0.9701      & \ding{55}    & \ding{55}   & \ding{55} \\
MAE/$_\text{LPIPS}$ & 1.74e-03   & 0.0375  & 0.9672   & \ding{55}    & \ding{55}   & \ding{55} \\
\cellcolor{lightcyan} OWT    & 8.66e-04   & 0.0248  & 0.9851   & 5.29e-04   & 0.0089   & 0.9759    \\
\cellcolor{lightcyan} OWT/$_{\text{4-slice}}$    & \textbf{7.07e-04}   & \textbf{0.0218}  & \textbf{0.9876}   & \textbf{4.47e-04}   & \textbf{0.0082}   & \textbf{0.9801}    \\
\bottomrule
\end{tabular}
}
\captionof{table}{Performance of image reconstruction on MRI datasets.}
\label{tab: recon_mr}
\end{minipage}
\end{table*}

\subsection{Advances in Design and Architecture}
\label{subsec:Advances}

Due to our proposed TGR training paradigm, the OWT framework produces semantically disentangled token groups (SDTGs) rich in organ-specific information. Leveraging the AHER module, these SDTGs can be individually or flexibly combined and restored from semantic embeddings into their original spatial dimensions. Consequently, as shown in Table~\ref{tab: downcheck}, OWT requires \textbf{only one-time training to enable multiple downstream tasks}. This cross-task efficiency and controllability significantly distinguish OWT from conventional networks with entangled embeddings, which cannot achieve such flexibility.

OWT is also computationally efficient. As shown in Table~\ref{tab: calculate}, OWT requires only 24.01, notably fewer than MAE’s 27.77 GFLOPs. This efficiency stems from (1) OWT's semantic tokens containing more highly condensed information than the traditional entangled holistic tokens, and (2) the lightweight design of the Organ Collector (0.25 GFLOPs) and AHER (0.074 GFLOPs) modules.

These designs also offer substantial potential. Specifically, we explore adaptively distributing different token numbers to SDTGs (\textit{e.g.,} based on organ volumes) in \textit{Appendix~\ref{sec: distribute}}. Since Organ Collector and AHER are encoder-decoder independent, we further investigate notable performance gains by replacing Transformer-based structures with VQGAN-based alternatives in \textit{Appendix~\ref{sec: BAlteration}}.

\begin{figure}[t]
\centering
\includegraphics[width=0.50\textwidth]{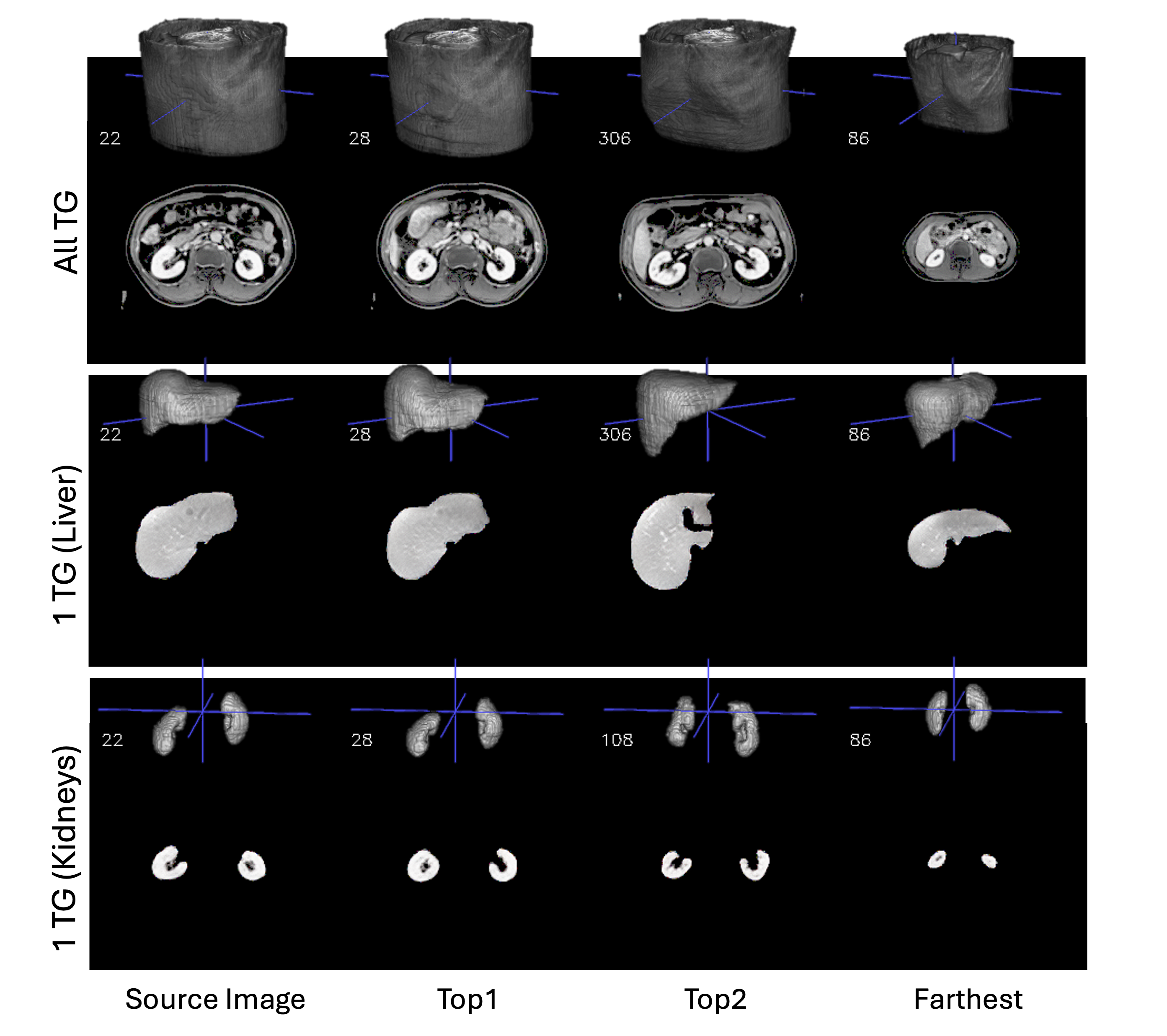}
\caption{Semantic-based organ-level retrieval of MRI images.} 
\label{fig: medret}
\end{figure}

%-----------------------------------------------------------------------------------------------------------------------------------
%-----------------------------------------------------------------------------------------------------------------------------------

\subsection{Unique Clinical Impacts}
\label{subsec:clinical}

\noindent
\textbf{Organ-Specific Tumor Identification.} In contrast to conventional models with entangled holistic embeddings, OWT extracts organ-specific information into distinct, semantically disentangled token groups (SDTGs). This structure enables the efficient use of only relevant tokens for highly specific downstream tasks. We demonstrate this on a classification task derived from tumor labels in the Abdomen1k dataset. As shown in Table~\ref{tab: tumor}, OWT leverages only 20 organ-specific tokens (\ie, for liver or kidneys) yet achieves significantly better performance than the holistic embedding-based MAE, which utilizes 196 tokens. This results in accuracy improvements of 6.2\% (79.7 vs. 73.5) and 7.1\% (81.8 vs. 74.7) for identifying tumors in the liver and kidneys, respectively, highlighting OWT's ability to create condensed yet highly informative representations for targeted clinical tasks.

\noindent
\textbf{Organ-Level Retrieval.}
Due to their semantic independence, our token groups enable organ-level retrieval ($L_2$ distance measure), a capability holistic embedding-based models cannot perform (Fig.~\ref{fig: medret}). We demonstrate top-1, top-2, and maximum-distance retrieval results on the MRI modality (CT examples are in \textit{Appendix~\ref{sec: retrict}}), where case indices in the top-left corner show results for different organs of interest. Since each token group preserves its organ's spatial structure and texture details, this targeted retrieval is highly beneficial in medical scenarios. For example, it can help radiologists quickly locate similar organ regions across patients without being influenced by other organs, facilitate region-based pathology comparisons, or assist in selecting reference images for organ-specific diagnoses.

%-----------------------------------------------------------------------------------------------------------------------------------
%-----------------------------------------------------------------------------------------------------------------------------------

\subsection{Downstream Analysis}
\label{subsec:down}
%-----------------------------------------------------------------------------------------------------------------------------------
%-----------------------------------------------------------------------------------------------------------------------------------

%-----------------------------------------------------------------------------------------------------------------------------------
%-----------------------------------------------------------------------------------------------------------------------------------
\noindent

%-----------------------------------------------------------------------------------------------------------------------------------
%-----------------------------------------------------------------------------------------------------------------------------------

\noindent
\textbf{Image Reconstruction.}
We evaluate OWT's reconstruction capability on four datasets (two CT, two MRI)~\cite{ma2021abdomenct, Gatidis2022, luo2024rethinking}. Tables~\ref{tab: recon_ct} and \ref{tab: recon_mr} show OWT outperforms MAE on L2, LPIPS, and 3D SSIM metrics in holistic reconstruction. Moreover, leveraging a 3D kernel for patch embedding in the encoder to tokenize four-slice 3D volumes (OWT/$_\text{4-slice}$) generates better results than the 2D OWT variant.

We further explore semantic-based reconstruction by retaining a single organ-wise token group, which is fed through AHER and the decoder to generate an organ-specific image. As reported in Tables~\ref{tab: recon_ct} and \ref{tab: recon_mr}, using more slices leads to more robust token groups and improves semantic reconstruction quality. Fig.~\ref{FigRec1} illustrates this on MRI cases: even with LPIPS added, MAE suffers from noticeable detail loss (red arrows), whereas OWT (all token groups retained) produces sharper, clearer details. Additionally, token groups can be freely combined to reconstruct only content of interest.

The zoom-in rows of Fig.~\ref{FigRec1} show that the liver/kidney reconstruction appears nearly identical whether using only its specific token group or combining it with others, suggesting OWT’s organ token groups are highly semantically independent. This property enhances interpretability and controllability, enabling reconstruction without unnecessary entanglement of unrelated organ features.

%-----------------------------------------------------------------------------------------------------------------------------------
%-----------------------------------------------------------------------------------------------------------------------------------
\begin{table}[t]  
\centering
\resizebox{1.0\columnwidth}{!}{
\begin{tabular}{lcccc}
\toprule
\multirow{2}{*}{1- / 4-slice} & \multicolumn{2}{c}{CT} & \multicolumn{2}{c}{MRI} \\
\cmidrule(lr){2-3} \cmidrule(lr){4-5}
 & Abdmen1k & AutoPet & Delay & PreArtery \\
\midrule
UNetplus & 88.38 / 89.78 & 89.29 / 91.04 & 84.11 / 85.50 & 84.42 / 85.68 \\
AttU\_Net & 89.34 / 90.75  & 89.73 / 91.46 & 84.54 / 86.75 & 85.04 / 86.52 \\
TransUnet & 88.91 / 90.24 & 89.70 / 91.34 & 84.97 / 86.40 & 84.29 / 86.01 \\
Swin-Unet & 86.56 / 89.47 & 88.06 / 89.91 & 83.07 / 85.39 & 82.41 / 84.48 \\
CMUNet & 88.33 / 90.92 & 89.48 / \textbf{91.50} & 84.82 / 86.82 & 85.56 / 86.94 \\
\midrule
\cellcolor{lightcyan} OWT  & 89.30 / \textbf{91.37} & 89.52 / 91.41 & 85.85 / \textbf{87.72} & 85.67 / \textbf{87.31} \\
\bottomrule
\end{tabular}
}
\caption{Dice performance (1- / 4-slice) of semantic segmentation.}
\label{tab: seg_ct1}
\end{table}

\begin{table}[t]
\centering
\resizebox{0.95\columnwidth}{!}{
\begin{tabular}{lccccc}
\toprule
\multirow{2}{*}{AutoPet} & \multicolumn{2}{c}{Holistic Rec.} & \multicolumn{2}{c}{1 TG Rec.} & \multicolumn{1}{c}{1 TG Seg.} \\
\cmidrule(lr){2-3} \cmidrule(lr){4-5} \cmidrule(lr){6-6}
& L2$\downarrow$ & SSIM$\uparrow$ & L2$\downarrow$ & SSIM$\uparrow$ & Dice$\uparrow$ \\
\midrule
MAE/$_\text{LPIPS}$  & 3.58e-04 & 0.9872& \ding{55} & \ding{55} & \ding{55} \\
AttU\_Net & \ding{55} & \ding{55} & \ding{55} & \ding{55} & 87.03 \\
CMUNet & \ding{55} & \ding{55} & \ding{55} & \ding{55} & 87.89 \\
\midrule
\cellcolor{lightcyan} OWT & 1.12e-04 & 0.9909 & 4.07e-04 & 0.9765 & 87.54 \\
\cellcolor{lightcyan} OWT/$_{\text{4-slice}}$ & \textbf{9.34e-05} & \textbf{0.9924} & \textbf{3.60e-04} & \textbf{0.9798} & \textbf{89.42} \\
\bottomrule
\end{tabular}
}
\caption{Generalization capability of cross-dataset experiments trained on Abdomen1k and tested on AutoPet.}
\label{tab: cross}
\end{table}

\noindent
\textbf{Image Segmentation.} Reconstructing OWT's token groups preserves clear spatial details and boundaries (Fig.~\ref{FigRec1}). Consequently, a simple thresholding operation on the predictions can generate semantic binary masks, offering direct evidence of each organ-wise token group’s independence and interpretability (More details on the thresholding operation are in \textit{Appendix~\ref{subsec: exset}}). Table~\ref{tab: seg_ct1} reports the single-organ segmentation Dice scores (averaged over the liver, kidneys, spleen, and pancreas). For fair comparison, all networks report segmentation results based on 1 and 4 slices. Overall, the accuracy of OWT incorporating information from adjacent slices (OWT with 4-slice) achieves comparable or superior performance across CT and MRI datasets. Detailed per-organ Dice comparisons are in \textit{Appendix~\ref{sec: seg}}.

%-----------------------------------------------------------------------------------------------------------------------------------
%-----------------------------------------------------------------------------------------------------------------------------------
\noindent
\textbf{Generalization.} Table~\ref{tab: cross} reports the generalization performance when all methods are trained on Abdomen1k and tested on the AutoPet dataset. Under both reconstruction and segmentation metrics, OWT/$_{\text{4-slice}}$ outperforms holistic-based approaches. It also achieves strong results in single-organ reconstruction and segmentation, indicating that token groups can be flexibly combined to mitigate interference from irrelevant organs and background regions, which is what traditional dedicated reconstruction and segmentation networks cannot do, underscoring OTW's potential as a novel foundational tokenization framework.

%-----------------------------------------------------------------------------------------------------------------------------------
%-----------------------------------------------------------------------------------------------------------------------------------

% \noindent
\subsection{Semantic Information in Token Groups}
\begin{figure}[t]
\centering
\includegraphics[width=0.49\textwidth]{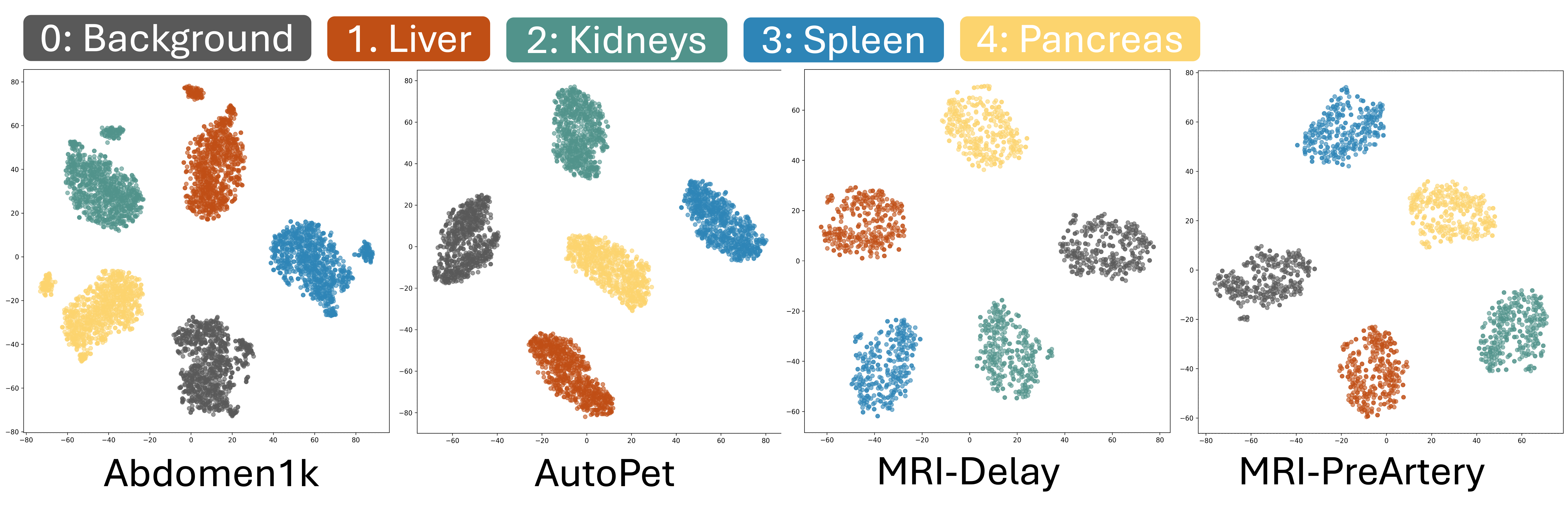}
\caption{t-SNE visualization of independent token groups on four medical datasets.} 
\label{fig: tsne4}
\end{figure}

\begin{table}[t]
\centering
\resizebox{0.9\columnwidth}{!}{
\begin{tabular}{lcccc}
\toprule
\multirow{2}{*}{ AutoPet} & \multicolumn{2}{c}{Holistic Rec.} & \multicolumn{2}{c}{1 TG Rec.} \\
\cmidrule(lr){2-3} \cmidrule(lr){4-5}
& L2 $\downarrow$ & LPIPS $\downarrow$ & L2 $\downarrow$ & LPIPS $\downarrow$ \\
\midrule
OWT & 1.49e-04 & 0.0232 & 3.32e-04 & 0.0071 \\
OWT \textit{w/} CT-CLIP & \textbf{1.39e-04}  & \textbf{0.0211} & \textbf{3.20e-04} & \textbf{0.0068} \\
\bottomrule
\end{tabular}
}
\caption{Reconstruction comparison between OWT \textit{w/} and \textit{w/o} incorporated text prompts.}
\label{tab: text}
\end{table}

\begin{figure}[t]
\centering
\includegraphics[width=0.49\textwidth]{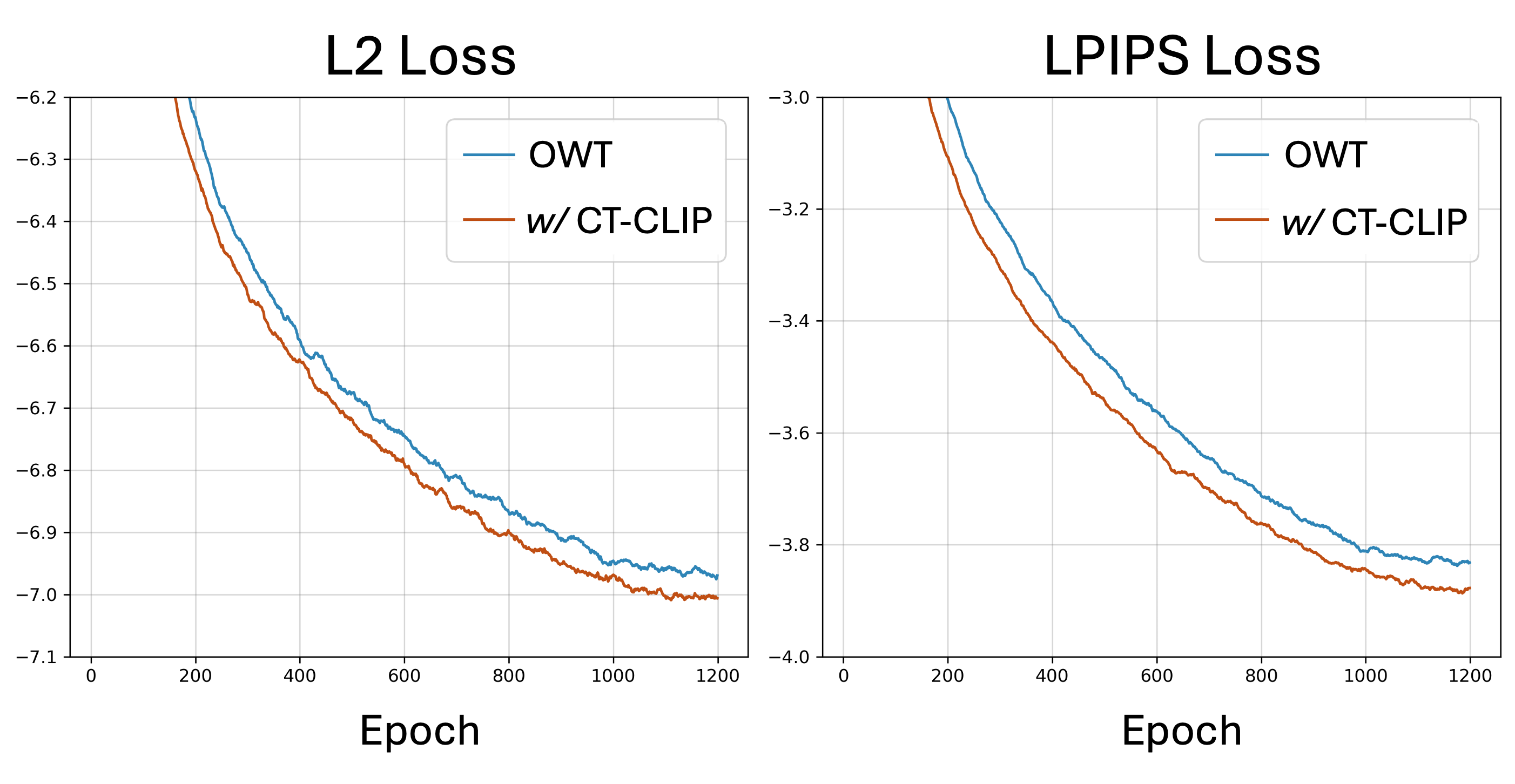}
\caption{Loss comparison between OWT \textit{w/} and \textit{w/o} incorporated text prompts.} 
\label{fig: text}
\end{figure}

\label{subsec:look}
\noindent
\textbf{T-SNE Analysis.}
Fig.~\ref{fig: tsne4} illustrates the t-SNE results of how the Organ Collector (with TGR) embeds the background and four organs into the latent space (\textit{i.e.}, five Token Groups) for each dataset. Within this space, token vectors of the same group type exhibit dense clusters, whereas different token groups remain distinctly separated. This clear inter-group separation highlights the semantic independence of organ-specific token groups in OWT.

\noindent
\textbf{Multi-Modality Incorporation.} OWT's highly disentangled and independent SDTGs enable efficient alignment or incorporation with other modalities, such as text. We demonstrate this by encoding medical prompts with a CT-CLIP encoder (specifically pretrained on paired CT images and reports)~\cite{hamamci2024developing}, and adding the resulting organ-specific and CT-consistent text embeddings element-wise to their corresponding token groups. More details on prompts and architecture for training OWT with text encoding are in \textit{Appendix~\ref{sec: multimodal}}.

As shown in Fig.~\ref{fig: text} and Table~\ref{tab: text}, this simple integration accelerates loss convergence and provides immediate improvements in several downstream tasks. This highlights OWT's promise for multi-modal enhancement, unlocking potential applications, such as text-guided retrieval, generation, and improved interpretability in complex workflows.

%-----------------------------------------------------------------------------------------------------------------------------------
%-----------------------------------------------------------------------------------------------------------------------------------

\begin{table}[t]
\centering
\resizebox{0.95\columnwidth}{!}{
\begin{tabular}{lcccccc}
\toprule
\multirow{2}{*}{AutoPet} & Data & \multicolumn{2}{c}{Holistic Rec.} & \multicolumn{2}{c}{1 TG Rec.} & Seg. \\
\cmidrule(lr){3-4} \cmidrule(lr){5-6}
 & Percent & L2$\downarrow$ &LPIPS $\downarrow$ & L2$\downarrow$ & LPIPS$\downarrow$ & Dice$\uparrow$ \\
\midrule
OWT & 100\% & 1.49e-04 & 0.0232 & 3.32e-04 & 0.0071 & 0.8952 \\
\midrule
OWT & 20\% & 5.54e-04 & 0.0658 & 5.70e-04 & 0.0115 & 0.8109 \\
OWT/$_\text{semi}$ & 20\% & \textbf{2.83e-04} & \textbf{0.0394} & \textbf{4.84e-04} & \textbf{0.0102} & \textbf{0.8415} \\
\midrule
OWT & 60\% & 1.86e-04 & 0.0270 & 3.49e-04 & 0.0076 & 0.8858 \\
OWT/$_\text{semi}$ & 60\% & \textbf{1.60e-04} & \textbf{0.0243} & \textbf{3.35e-04} & \textbf{0.0072} & \textbf{0.8907} \\
\bottomrule
\end{tabular}
}
\caption{Performance comparison of OWT in different training strategies.}
\label{tab: semisuper}
\end{table}

\begin{table}[t]
\centering
\resizebox{0.8\columnwidth}{!}{
\begin{tabular}{lcccc}
\toprule
\multirow{2}{*}{CelebAMaskHQ} & \multicolumn{2}{c}{Holistic Rec.} & \multicolumn{2}{c}{1 Token Group Rec.} \\
\cmidrule(lr){2-3} \cmidrule(lr){4-5}
& L2 $\downarrow$ & LPIPS $\downarrow$ & L2 $\downarrow$ & LPIPS $\downarrow$ \\
\midrule
MAE/$_\text{LPIPS}$ & 2.78e-03 & 0.1157 & \ding{55} & \ding{55} \\
\cellcolor{lightcyan} OWT & \textbf{2.50e-03} & \textbf{0.1101} & {2.83e-03} & {0.0458}  \\
\bottomrule
\end{tabular}
}
\caption{Performance of OWT on CelebAMaskHQ dataset.}
\label{tab: facerec}
\end{table}

\begin{figure}[t]
\centering
\includegraphics[width=0.45\textwidth]{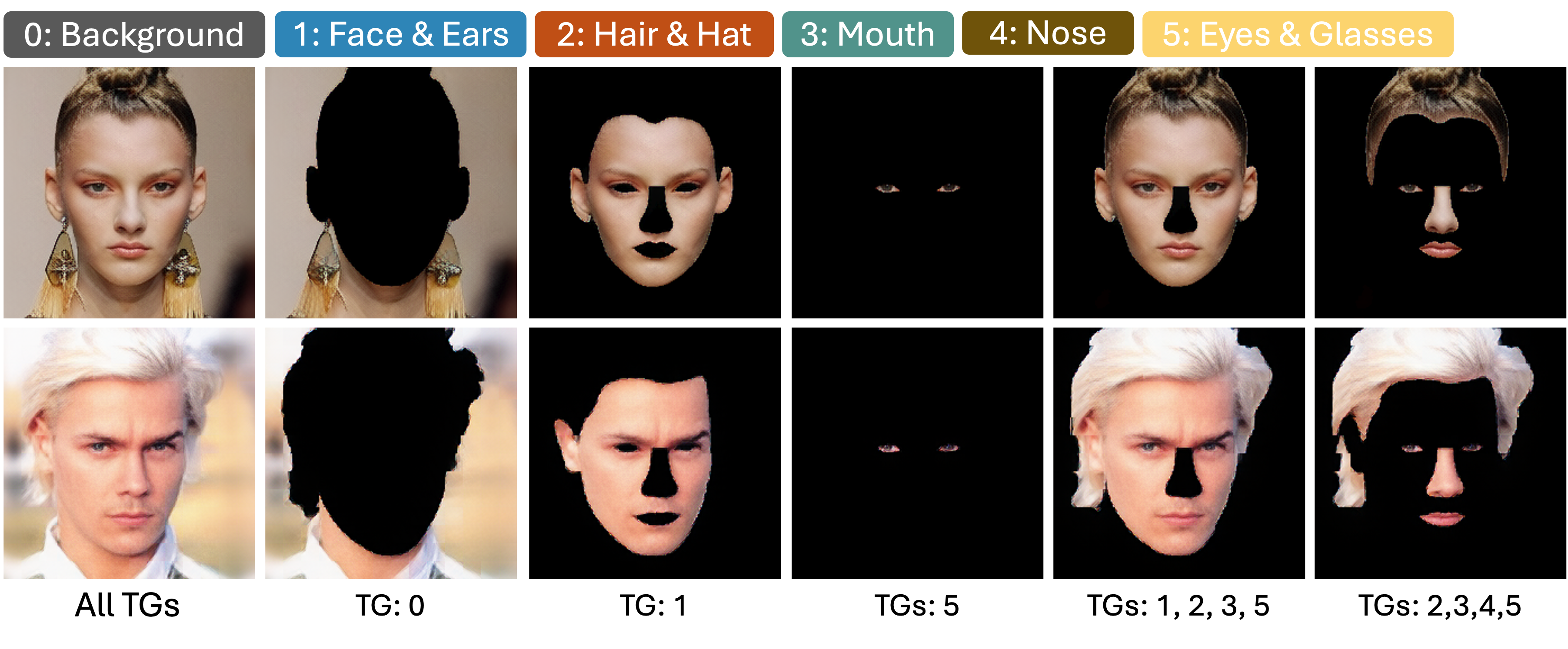}
\caption{Samples of holistic and semantic-based human face reconstruction on CelebAMaskHQ.} 
\label{fig: facegen}
\end{figure}

\subsection{Semi-Supervised Training under Limited Annotations} 
\label{sec: percelabel}

The limitation of annotation is a long-standing issue in medical imaging. In our foundational framework, we find that appropriate supervision of token groups is beneficial for downstream tasks. Therefore, we adopted a two-step training strategy: (1) using all data \textit{without} labels for unsupervised pretraining of OWT’s encoder and decoder via whole image reconstruction for 600 epochs, and (2) fine-tuning the entire OWT with only \textit{partial} annotation coverage (20\% and 60\%) from the training set for 1200 epochs. 

As shown in Table~\ref{tab: semisuper}, this approach significantly improves performance under label scarcity. The semi-supervised model using 20\% labels greatly outperformed the model trained on only 20\% labels. Furthermore, the semi-supervised mode with 60\% labels nearly matched the upper-bound results of training with 100\% labels , showing OWT's robustness and potential for data-scarce scenarios.

%-----------------------------------------------------------------------------------------------------------------------------------
%-----------------------------------------------------------------------------------------------------------------------------------

\subsection{Token Group as Compositional Units}
\label{subsec:face}

\begin{figure}[t]
\centering
\includegraphics[width=0.49\textwidth]{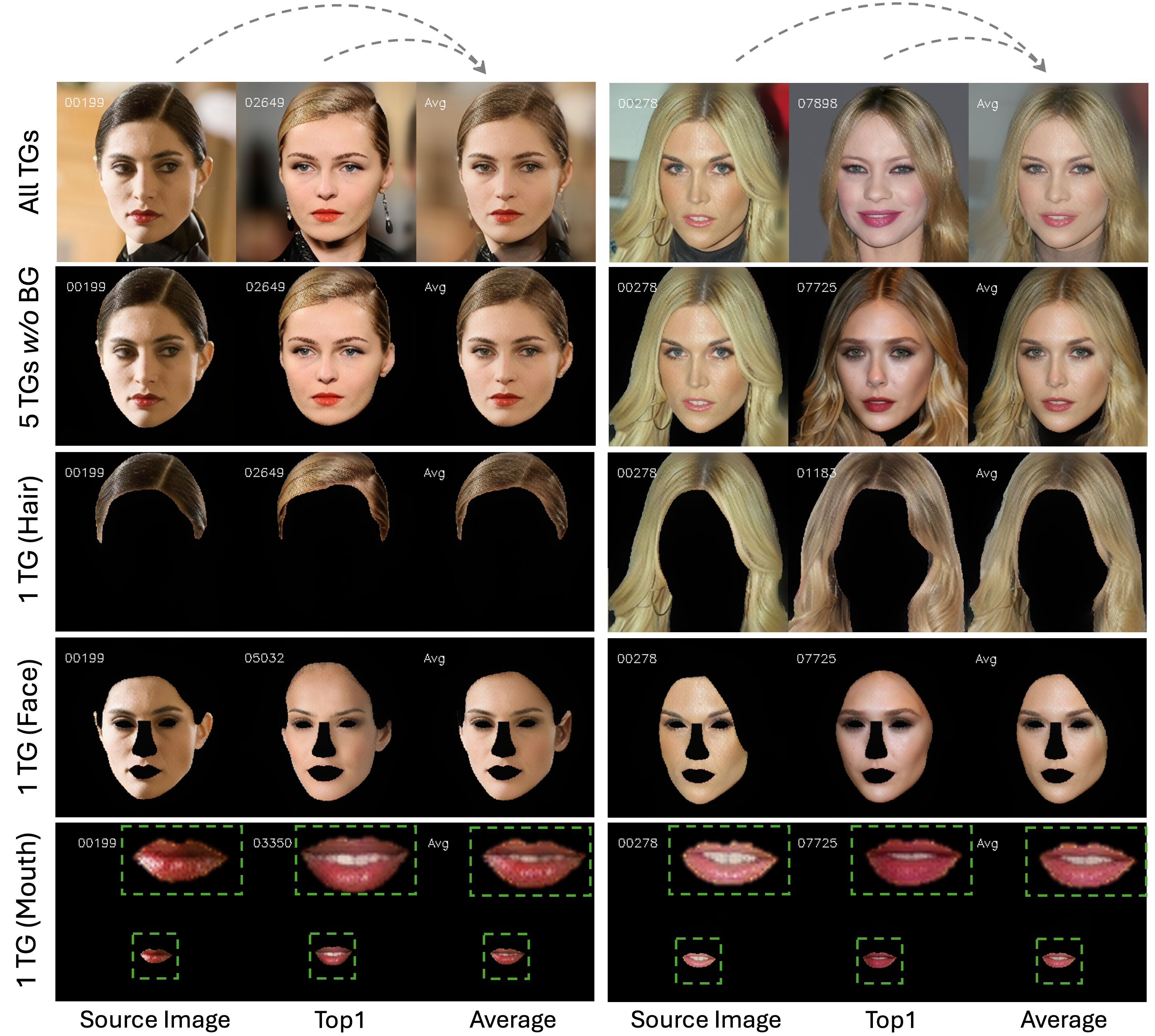}
\caption{Semantic-based retrieval of CelebAMaskHQ.} 
\label{fig: faceret}
\end{figure}
Although OWT and TGR are primarily designed for grouping different semantic components in medical images, they can also be applied to non-medical datasets. As a proof of concept, we demonstrate in Fig.~\ref{fig: facegen} and Table~\ref{tab: facerec} that OWT successfully learns semantically disentangled token groups on the CelebAMaskHQ dataset, capturing and reconstructing each facial region with high spatial and textural fidelity.

As illustrated in Fig.~\ref{fig: faceret}, each token group facilitates semantic-based retrieval for different facial attributes (\textit{e.g.}, hair, skin, lips). In the third column, we conduct an experiment by averaging source token group(s) with the Top1 retrieved token group(s). Notably, this averaged token group(s) can still be reconstructed, exhibiting features that blend and transition between both identities (\textit{e.g.}, overall facial attributes, hair color, skin tone, lip color, mouth openness). These observations imply that token groups extracted by OWT can enable fine-grained editing or hybrid generation of facial attributes without background influence, offering enhanced interpretability, efficiency, and controllability.

%%%%%%%%%%%%%%%%%%%%%%%%%%%%%%%%%%%%%%%%%%%%%%%%%%%%%%%%%%%%%%%%%%%%%%%%%%%%%%%%%%%%%
%%%%%%%%%%%%%%%%%%%%%%%%%%%%%%%%%%%%%%%%%%%%%%%%%%%%%%%%%%%%%%%%%%%%%%%%%%%%%%%%%%%%%
\section{Conclusion}
In this paper, we propose a novel Organ-Wise Tokenization (OWT) framework and Token Group-Based Reconstruction (TGR) paradigm, which disentangle holistic embeddings into semantically disentangled token groups. Our approach not only improves interpretability, generalizability, efficiency and controllability in organ-based analyses, but also shows adaptability to broader scenarios, including natural image domains. By enabling semantic-level token groups, OWT enriches unique clinical applications and downstream tasks such as organ-specific analysis, semantic-based reconstruction, segmentation and retrieval, offering a new perspective on how representations can be utilized.
{
    \small
    \bibliographystyle{ieeenat_fullname}
    \bibliography{main}

@article{bank2023autoencoders,
  title={Autoencoders},
  author={Bank, Dor and Koenigstein, Noam and Giryes, Raja},
  journal={Machine learning for data science handbook: data mining and knowledge discovery handbook},
  pages={353--374},
  year={2023},
  publisher={Springer}
}

@inproceedings{he2022masked,
  title={Masked autoencoders are scalable vision learners},
  author={He, Kaiming and Chen, Xinlei and Xie, Saining and Li, Yanghao and Doll{\'a}r, Piotr and Girshick, Ross},
  booktitle={Proceedings of the IEEE/CVF conference on computer vision and pattern recognition},
  pages={16000--16009},
  year={2022}
}

@article{feichtenhofer2022masked,
  title={Masked autoencoders as spatiotemporal learners},
  author={Feichtenhofer, Christoph and Li, Yanghao and He, Kaiming and others},
  journal={Advances in neural information processing systems},
  volume={35},
  pages={35946--35958},
  year={2022}
}

@article{dosovitskiy2020image,
  title={An image is worth 16x16 words: Transformers for image recognition at scale},
  author={Dosovitskiy, Alexey and Beyer, Lucas and Kolesnikov, Alexander and Weissenborn, Dirk and Zhai, Xiaohua and Unterthiner, Thomas and Dehghani, Mostafa and Minderer, Matthias and Heigold, Georg and Gelly, Sylvain and others},
  journal={arXiv preprint arXiv:2010.11929},
  year={2020}
}

@inproceedings{esser2021taming,
  title={Taming transformers for high-resolution image synthesis},
  author={Esser, Patrick and Rombach, Robin and Ommer, Bjorn},
  booktitle={Proceedings of the IEEE/CVF conference on computer vision and pattern recognition},
  pages={12873--12883},
  year={2021}
}

@inproceedings{zhang2018unreasonable,
  title={The unreasonable effectiveness of deep features as a perceptual metric},
  author={Zhang, Richard and Isola, Phillip and Efros, Alexei A and Shechtman, Eli and Wang, Oliver},
  booktitle={Proceedings of the IEEE conference on computer vision and pattern recognition},
  pages={586--595},
  year={2018}
}

@inproceedings{katharopoulos2020transformers,
  title={Transformers are rnns: Fast autoregressive transformers with linear attention},
  author={Katharopoulos, Angelos and Vyas, Apoorv and Pappas, Nikolaos and Fleuret, Fran{\c{c}}ois},
  booktitle={International conference on machine learning},
  pages={5156--5165},
  year={2020},
  organization={PMLR}
}

@inproceedings{zhou2018unet++,
  title={Unet++: A nested u-net architecture for medical image segmentation},
  author={Zhou, Zongwei and Rahman Siddiquee, Md Mahfuzur and Tajbakhsh, Nima and Liang, Jianming},
  booktitle={Deep learning in medical image analysis and multimodal learning for clinical decision support: 4th international workshop, DLMIA 2018, and 8th international workshop, ML-CDS 2018, held in conjunction with MICCAI 2018, Granada, Spain, September 20, 2018, proceedings 4},
  pages={3--11},
  year={2018},
  organization={Springer}
}

@article{oktay2018attention,
  title={Attention u-net: Learning where to look for the pancreas},
  author={Oktay, Ozan and Schlemper, Jo and Folgoc, Loic Le and Lee, Matthew and Heinrich, Mattias and Misawa, Kazunari and Mori, Kensaku and McDonagh, Steven and Hammerla, Nils Y and Kainz, Bernhard and others},
  journal={arXiv preprint arXiv:1804.03999},
  year={2018}
}

@article{chen2021transunet,
  title={Transunet: Transformers make strong encoders for medical image segmentation},
  author={Chen, Jieneng and Lu, Yongyi and Yu, Qihang and Luo, Xiangde and Adeli, Ehsan and Wang, Yan and Lu, Le and Yuille, Alan L and Zhou, Yuyin},
  journal={arXiv preprint arXiv:2102.04306},
  year={2021}
}

@inproceedings{cao2022swin,
  title={Swin-unet: Unet-like pure transformer for medical image segmentation},
  author={Cao, Hu and Wang, Yueyue and Chen, Joy and Jiang, Dongsheng and Zhang, Xiaopeng and Tian, Qi and Wang, Manning},
  booktitle={European conference on computer vision},
  pages={205--218},
  year={2022},
  organization={Springer}
}

@inproceedings{tang2023cmu,
  title={Cmu-net: a strong convmixer-based medical ultrasound image segmentation network},
  author={Tang, Fenghe and Wang, Lingtao and Ning, Chunping and Xian, Min and Ding, Jianrui},
  booktitle={2023 IEEE 20th international symposium on biomedical imaging (ISBI)},
  pages={1--5},
  year={2023},
  organization={IEEE}
}

@article{wang2024disentangled,
  title={Disentangled representation learning},
  author={Wang, Xin and Chen, Hong and Wu, Zihao and Zhu, Wenwu and others},
  journal={IEEE Transactions on Pattern Analysis and Machine Intelligence},
  year={2024},
  publisher={IEEE}
}

@article{geirhos2020shortcut,
  title={Shortcut learning in deep neural networks},
  author={Geirhos, Robert and Jacobsen, J{\"o}rn-Henrik and Michaelis, Claudio and Zemel, Richard and Brendel, Wieland and Bethge, Matthias and Wichmann, Felix A},
  journal={Nature Machine Intelligence},
  volume={2},
  number={11},
  pages={665--673},
  year={2020},
  publisher={Nature Publishing Group UK London}
}

@article{voulodimos2018deep,
  title={Deep learning for computer vision: A brief review},
  author={Voulodimos, Athanasios and Doulamis, Nikolaos and Doulamis, Anastasios and Protopapadakis, Eftychios},
  journal={Computational intelligence and neuroscience},
  volume={2018},
  number={1},
  pages={7068349},
  year={2018},
  publisher={Wiley Online Library}
}

@article{gidaris2018unsupervised,
  title={Unsupervised representation learning by predicting image rotations},
  author={Gidaris, Spyros and Singh, Praveer and Komodakis, Nikos},
  journal={arXiv preprint arXiv:1803.07728},
  year={2018}
}

@article{oord2018representation,
  title={Representation learning with contrastive predictive coding},
  author={Oord, Aaron van den and Li, Yazhe and Vinyals, Oriol},
  journal={arXiv preprint arXiv:1807.03748},
  year={2018}
}

@inproceedings{chen2020simple,
  title={A simple framework for contrastive learning of visual representations},
  author={Chen, Ting and Kornblith, Simon and Norouzi, Mohammad and Hinton, Geoffrey},
  booktitle={International conference on machine learning},
  pages={1597--1607},
  year={2020},
  organization={PmLR}
}

@inproceedings{chen2021exploring,
  title={Exploring simple siamese representation learning},
  author={Chen, Xinlei and He, Kaiming},
  booktitle={Proceedings of the IEEE/CVF conference on computer vision and pattern recognition},
  pages={15750--15758},
  year={2021}
}

@article{grill2020bootstrap,
  title={Bootstrap your own latent-a new approach to self-supervised learning},
  author={Grill, Jean-Bastien and Strub, Florian and Altch{\'e}, Florent and Tallec, Corentin and Richemond, Pierre and Buchatskaya, Elena and Doersch, Carl and Avila Pires, Bernardo and Guo, Zhaohan and Gheshlaghi Azar, Mohammad and others},
  journal={Advances in neural information processing systems},
  volume={33},
  pages={21271--21284},
  year={2020}
}

@article{bao2021beit,
  title={Beit: Bert pre-training of image transformers},
  author={Bao, Hangbo and Dong, Li and Piao, Songhao and Wei, Furu},
  journal={arXiv preprint arXiv:2106.08254},
  year={2021}
}

@inproceedings{he2016deep,
  title={Deep residual learning for image recognition},
  author={He, Kaiming and Zhang, Xiangyu and Ren, Shaoqing and Sun, Jian},
  booktitle={Proceedings of the IEEE conference on computer vision and pattern recognition},
  pages={770--778},
  year={2016}
}

@inproceedings{huang2017densely,
  title={Densely connected convolutional networks},
  author={Huang, Gao and Liu, Zhuang and Van Der Maaten, Laurens and Weinberger, Kilian Q},
  booktitle={Proceedings of the IEEE conference on computer vision and pattern recognition},
  pages={4700--4708},
  year={2017}
}

@article{chen2021momentum,
  title={Momentum contrastive learning for few-shot COVID-19 diagnosis from chest CT images},
  author={Chen, Xiaocong and Yao, Lina and Zhou, Tao and Dong, Jinming and Zhang, Yu},
  journal={Pattern recognition},
  volume={113},
  pages={107826},
  year={2021},
  publisher={Elsevier}
}

@article{lai2024e3d,
  title={E3D-GPT: Enhanced 3D Visual Foundation for Medical Vision-Language Model},
  author={Lai, Haoran and Jiang, Zihang and Yao, Qingsong and Wang, Rongsheng and He, Zhiyang and Tao, Xiaodong and Wei, Wei and Lv, Weifu and Zhou, S Kevin},
  journal={arXiv preprint arXiv:2410.14200},
  year={2024}
}

@article{chaitanya2020contrastive,
  title={Contrastive learning of global and local features for medical image segmentation with limited annotations},
  author={Chaitanya, Krishna and Erdil, Ertunc and Karani, Neerav and Konukoglu, Ender},
  journal={Advances in neural information processing systems},
  volume={33},
  pages={12546--12558},
  year={2020}
}

@article{lang20233d,
  title={3d masked autoencoders with application to anomaly detection in non-contrast enhanced breast mri},
  author={Lang, Daniel M and Schwartz, Eli and Bercea, Cosmin I and Giryes, Raja and Schnabel, Julia A},
  journal={arXiv preprint arXiv:2303.05861},
  year={2023}
}

@article{tiu2022expert,
  title={Expert-level detection of pathologies from unannotated chest X-ray images via self-supervised learning},
  author={Tiu, Ekin and Talius, Ellie and Patel, Pujan and Langlotz, Curtis P and Ng, Andrew Y and Rajpurkar, Pranav},
  journal={Nature biomedical engineering},
  volume={6},
  number={12},
  pages={1399--1406},
  year={2022},
  publisher={Nature Publishing Group UK London}
}

@inproceedings{xiao2023delving,
  title={Delving into masked autoencoders for multi-label thorax disease classification},
  author={Xiao, Junfei and Bai, Yutong and Yuille, Alan and Zhou, Zongwei},
  booktitle={Proceedings of the IEEE/CVF Winter Conference on Applications of Computer Vision},
  pages={3588--3600},
  year={2023}
}

@inproceedings{ronneberger2015u,
  title={U-net: Convolutional networks for biomedical image segmentation},
  author={Ronneberger, Olaf and Fischer, Philipp and Brox, Thomas},
  booktitle={Medical image computing and computer-assisted intervention--MICCAI 2015: 18th international conference, Munich, Germany, October 5-9, 2015, proceedings, part III 18},
  pages={234--241},
  year={2015},
  organization={Springer}
}

@inproceedings{hatamizadeh2021swin,
  title={Swin unetr: Swin transformers for semantic segmentation of brain tumors in mri images},
  author={Hatamizadeh, Ali and Nath, Vishwesh and Tang, Yucheng and Yang, Dong and Roth, Holger R and Xu, Daguang},
  booktitle={International MICCAI brainlesion workshop},
  pages={272--284},
  year={2021},
  organization={Springer}
}

@article{schutte2021using,
  title={Using stylegan for visual interpretability of deep learning models on medical images},
  author={Schutte, Kathryn and Moindrot, Olivier and H{\'e}rent, Paul and Schiratti, Jean-Baptiste and J{\'e}gou, Simon},
  journal={arXiv preprint arXiv:2101.07563},
  year={2021}
}

@article{cetin2023attri,
  title={Attri-VAE: Attribute-based interpretable representations of medical images with variational autoencoders},
  author={Cetin, Irem and Stephens, Maialen and Camara, Oscar and Ballester, Miguel A Gonz{\'a}lez},
  journal={Computerized Medical Imaging and Graphics},
  volume={104},
  pages={102158},
  year={2023},
  publisher={Elsevier}
}

@article{dangovski2021equivariant,
  title={Equivariant contrastive learning},
  author={Dangovski, Rumen and Jing, Li and Loh, Charlotte and Han, Seungwook and Srivastava, Akash and Cheung, Brian and Agrawal, Pulkit and Solja{\v{c}}i{\'c}, Marin},
  journal={arXiv preprint arXiv:2111.00899},
  year={2021}
}

@article{bai2023robust,
  title={Robust and rotation-equivariant contrastive learning},
  author={Bai, Gairui and Xi, Wei and Hong, Xiaopeng and Liu, Xinhui and Yue, Yang and Zhao, Songwen},
  journal={IEEE Transactions on Neural Networks and Learning Systems},
  year={2023},
  publisher={IEEE}
}

@inproceedings{devillers2023equimod,
  title={Equimod: An equivariance module to improve visual instance discrimination},
  author={Devillers, Alexandre and Lefort, Mathieu},
  booktitle={International Conference on Learning Representations},
  year={2023}
}

@article{garrido2023self,
  title={Self-supervised learning of split invariant equivariant representations},
  author={Garrido, Quentin and Najman, Laurent and Lecun, Yann},
  journal={arXiv preprint arXiv:2302.10283},
  year={2023}
}

@inproceedings{wang2024distortion,
  title={Distortion-disentangled contrastive learning},
  author={Wang, Jinfeng and Song, Sifan and Su, Jionglong and Zhou, S Kevin},
  booktitle={Proceedings of the IEEE/CVF Winter Conference on Applications of Computer Vision},
  pages={75--85},
  year={2024}
}

@article{song2024contrastive,
  title={Contrastive Learning Via Equivariant Representation},
  author={Song, Sifan and Wang, Jinfeng and Zhao, Qiaochu and Li, Xiang and Wu, Dufan and Stefanidis, Angelos and Su, Jionglong and Zhou, S Kevin and Li, Quanzheng},
  journal={arXiv preprint arXiv:2406.00262},
  year={2024}
}

@article{burgess2018understanding,
  title={Understanding disentangling in beta-VAE},
  author={Burgess, Christopher P and Higgins, Irina and Pal, Arka and Matthey, Loic and Watters, Nick and Desjardins, Guillaume and Lerchner, Alexander},
  journal={arXiv preprint arXiv:1804.03599},
  year={2018}
}

@article{chen2016infogan,
  title={Infogan: Interpretable representation learning by information maximizing generative adversarial nets},
  author={Chen, Xi and Duan, Yan and Houthooft, Rein and Schulman, John and Sutskever, Ilya and Abbeel, Pieter},
  journal={Advances in neural information processing systems},
  volume={29},
  year={2016}
}

@article{liu2022learning,
  title={Learning disentangled representations in the imaging domain},
  author={Liu, Xiao and Sanchez, Pedro and Thermos, Spyridon and O’Neil, Alison Q and Tsaftaris, Sotirios A},
  journal={Medical Image Analysis},
  volume={80},
  pages={102516},
  year={2022},
  publisher={Elsevier}
}

@article{wang2022disentangled,
  title={Disentangled representation for cross-domain medical image segmentation},
  author={Wang, Jie and Zhong, Chaoliang and Feng, Cheng and Zhang, Ying and Sun, Jun and Yokota, Yasuto},
  journal={IEEE Transactions on Instrumentation and Measurement},
  volume={72},
  pages={1--15},
  year={2022},
  publisher={IEEE}
}

@article{wang2022semantic,
  title={Semantic-guided Disentangled Representation for Unsupervised Cross-modality Medical Image Segmentation},
  author={Wang, Shuai and Li, Rui},
  journal={arXiv preprint arXiv:2203.14025},
  year={2022}
}

@article{jiang2022disentangled,
  title={Disentangled representation and cross-modality image translation based unsupervised domain adaptation method for abdominal organ segmentation},
  author={Jiang, Kaida and Quan, Li and Gong, Tao},
  journal={International Journal of Computer Assisted Radiology and Surgery},
  volume={17},
  number={6},
  pages={1101--1113},
  year={2022},
  publisher={Springer}
}

@inproceedings{ouyang2021representation,
  title={Representation disentanglement for multi-modal brain MRI analysis},
  author={Ouyang, Jiahong and Adeli, Ehsan and Pohl, Kilian M and Zhao, Qingyu and Zaharchuk, Greg},
  booktitle={Information Processing in Medical Imaging: 27th International Conference, IPMI 2021, Virtual Event, June 28--June 30, 2021, Proceedings 27},
  pages={321--333},
  year={2021},
  organization={Springer}
}

@article{chartsias2019disentangled,
  title={Disentangled representation learning in cardiac image analysis},
  author={Chartsias, Agisilaos and Joyce, Thomas and Papanastasiou, Giorgos and Semple, Scott and Williams, Michelle and Newby, David E and Dharmakumar, Rohan and Tsaftaris, Sotirios A},
  journal={Medical image analysis},
  volume={58},
  pages={101535},
  year={2019},
  publisher={Elsevier}
}

@inproceedings{liu2022joint,
  title={Joint prediction of meningioma grade and brain invasion via task-aware contrastive learning},
  author={Liu, Tianling and Liu, Wennan and Yu, Lequan and Wan, Liang and Han, Tong and Zhu, Lei},
  booktitle={International Conference on Medical Image Computing and Computer-Assisted Intervention},
  pages={355--365},
  year={2022},
  organization={Springer}
}

@article{hu2022clinically,
  title={Clinically plausible pathology-anatomy disentanglement in patient brain mri with structured variational priors},
  author={Hu, Anjun and Falet, Jean-Pierre R and Nichyporuk, Brennan S and Shui, Changjian and Arnold, Douglas L and Tsaftaris, Sotirios A and Arbel, Tal},
  journal={arXiv preprint arXiv:2211.07820},
  year={2022}
}

@inproceedings{locatello2019challenging,
  title={Challenging common assumptions in the unsupervised learning of disentangled representations},
  author={Locatello, Francesco and Bauer, Stefan and Lucic, Mario and Raetsch, Gunnar and Gelly, Sylvain and Sch{\"o}lkopf, Bernhard and Bachem, Olivier},
  booktitle={international conference on machine learning},
  pages={4114--4124},
  year={2019},
  organization={PMLR}
}

@article{shen2017deep,
  title={Deep learning in medical image analysis},
  author={Shen, Dinggang and Wu, Guorong and Suk, Heung-Il},
  journal={Annual review of biomedical engineering},
  volume={19},
  number={1},
  pages={221--248},
  year={2017},
  publisher={Annual Reviews}
}

@article{mortada2023segmentation,
  title={Segmentation of anatomical structures of the left heart from echocardiographic images using deep learning},
  author={Mortada, MHD Jafar and Tomassini, Selene and Anbar, Haidar and Morettini, Micaela and Burattini, Laura and Sbrollini, Agnese},
  journal={Diagnostics},
  volume={13},
  number={10},
  pages={1683},
  year={2023},
  publisher={MDPI}
}

@article{weston2019automated,
  title={Automated abdominal segmentation of CT scans for body composition analysis using deep learning},
  author={Weston, Alexander D and Korfiatis, Panagiotis and Kline, Timothy L and Philbrick, Kenneth A and Kostandy, Petro and Sakinis, Tomas and Sugimoto, Motokazu and Takahashi, Naoki and Erickson, Bradley J},
  journal={Radiology},
  volume={290},
  number={3},
  pages={669--679},
  year={2019},
  publisher={Radiological Society of North America}
}

@article{xiao2017dna,
  title={Dna-gan: Learning disentangled representations from multi-attribute images},
  author={Xiao, Taihong and Hong, Jiapeng and Ma, Jinwen},
  journal={arXiv preprint arXiv:1711.05415},
  year={2017}
}

@article{zhu2018visual,
  title={Visual object networks: Image generation with disentangled 3D representations},
  author={Zhu, Jun-Yan and Zhang, Zhoutong and Zhang, Chengkai and Wu, Jiajun and Torralba, Antonio and Tenenbaum, Josh and Freeman, Bill},
  journal={Advances in neural information processing systems},
  volume={31},
  year={2018}
}

@article{ryoo2021tokenlearner,
  title={Tokenlearner: What can 8 learned tokens do for images and videos?},
  author={Ryoo, Michael S and Piergiovanni, AJ and Arnab, Anurag and Dehghani, Mostafa and Angelova, Anelia},
  journal={arXiv preprint arXiv:2106.11297},
  year={2021}
}

@inproceedings{fayyaz2022adaptive,
  title={Adaptive token sampling for efficient vision transformers},
  author={Fayyaz, Mohsen and Koohpayegani, Soroush Abbasi and Jafari, Farnoush Rezaei and Sengupta, Sunando and Joze, Hamid Reza Vaezi and Sommerlade, Eric and Pirsiavash, Hamed and Gall, J{\"u}rgen},
  booktitle={European Conference on Computer Vision},
  pages={396--414},
  year={2022},
  organization={Springer}
}

@article{ma2021abdomenct,
  title={Abdomenct-1k: Is abdominal organ segmentation a solved problem?},
  author={Ma, Jun and Zhang, Yao and Gu, Song and Zhu, Cheng and Ge, Cheng and Zhang, Yichi and An, Xingle and Wang, Congcong and Wang, Qiyuan and Liu, Xin and others},
  journal={IEEE Transactions on Pattern Analysis and Machine Intelligence},
  volume={44},
  number={10},
  pages={6695--6714},
  year={2021},
  publisher={IEEE}
}

@misc{Gatidis2022,
  author = {Gatidis, S. and Kuestner, T.},
  year = {2022},
  title = {A whole-body FDG-PET/CT dataset with manually annotated tumor lesions (FDG-PET-CT-Lesions)},
  howpublished = {Dataset},
  publisher = {The Cancer Imaging Archive},
  doi = {10.7937/gkr0-xv29}
}

@inproceedings{luo2024rethinking,
  title={Rethinking Abdominal Organ Segmentation (RAOS) in the clinical scenario: A robustness evaluation benchmark with challenging cases},
  author={Luo, Xiangde and Li, Zihan and Zhang, Shaoting and Liao, Wenjun and Wang, Guotai},
  booktitle={International Conference on Medical Image Computing and Computer-Assisted Intervention},
  pages={531--541},
  year={2024},
  organization={Springer}
}

@inproceedings{lee2020maskgan,
  title={Maskgan: Towards diverse and interactive facial image manipulation},
  author={Lee, Cheng-Han and Liu, Ziwei and Wu, Lingyun and Luo, Ping},
  booktitle={Proceedings of the IEEE/CVF conference on computer vision and pattern recognition},
  pages={5549--5558},
  year={2020}
}

@article{vaswani2017attention,
  title={Attention is all you need},
  author={Vaswani, Ashish and Shazeer, Noam and Parmar, Niki and Uszkoreit, Jakob and Jones, Llion and Gomez, Aidan N and Kaiser, {\L}ukasz and Polosukhin, Illia},
  journal={Advances in neural information processing systems},
  volume={30},
  year={2017}
}

@inproceedings{liu2023clip,
  title={Clip-driven universal model for organ segmentation and tumor detection},
  author={Liu, Jie and Zhang, Yixiao and Chen, Jie-Neng and Xiao, Junfei and Lu, Yongyi and A Landman, Bennett and Yuan, Yixuan and Yuille, Alan and Tang, Yucheng and Zhou, Zongwei},
  booktitle={Proceedings of the IEEE/CVF international conference on computer vision},
  pages={21152--21164},
  year={2023}
}

@misc{kingma2013auto,
  title={Auto-encoding variational bayes},
  author={Kingma, Diederik P and Welling, Max and others},
  year={2013},
  publisher={Banff, Canada}
}

@article{hamamci2024developing,
  title={Developing generalist foundation models from a multimodal dataset for 3d computed tomography},
  author={Hamamci, Ibrahim Ethem and Er, Sezgin and Wang, Chenyu and Almas, Furkan and Simsek, Ayse Gulnihan and Esirgun, Sevval Nil and Doga, Irem and Durugol, Omer Faruk and Dai, Weicheng and Xu, Murong and others},
  journal={arXiv preprint arXiv:2403.17834},
  year={2024}
}
}

\clearpage
\setcounter{page}{1}
\maketitlesupplementary

\appendix
\section{Related Work}
\label{sec:related}

\subsection{Holistic Embedding-based Representation Learning}

Holistic embedding-based approaches map entire 2D/3D inputs into a single latent representation, aiming to capture global image features~\cite{voulodimos2018deep, geirhos2020shortcut, wang2024disentangled}. These methods typically follow either {self-supervised} or {task-guided} strategies. In self-supervised learning, contrastive techniques like SimCLR~\cite{chen2020simple}, SimSiam~\cite{chen2021exploring}, and BYOL~\cite{grill2020bootstrap} enforce consistency between augmented views, while predictive approaches such as CPC~\cite{oord2018representation} and RotNet~\cite{gidaris2018unsupervised} reconstruct missing or permuted portions. Recently, masked image modeling (\textit{e.g.}, MAE~\cite{he2022masked} and BEiT~\cite{bao2021beit}) has emerged as an effective approach by requiring models to infer masked regions. Meanwhile, task-guided frameworks focus on extracting features that directly benefit a target task, typically resulting in holistic embeddings. Many classic neural networks, such as ResNet~\cite{he2016deep}, DenseNet~\cite{huang2017densely}, and Vision Transformer~\cite{dosovitskiy2020image}, and their variants incorporate classification, regression, or segmentation objectives.

In medical imaging, holistic representations have been widely used for self-supervised feature learning in CT~\cite{chen2021momentum, lai2024e3d}, MRI~\cite{chaitanya2020contrastive, lang20233d}, and X-ray~\cite{tiu2022expert, xiao2023delving}. Additionally, classical supervised pipelines (\textit{e.g.}, U-Net~\cite{ronneberger2015u} and its variants~\cite{zhou2018unet++, chen2021transunet, cao2022swin, hatamizadeh2021swin}) rely on a global bottleneck to embed entire images or volumes, demonstrating strong performance in organ segmentation and disease detection. Although such approaches achieve notable success, they often encode multiple semantic factors, such as multiple organs and the surrounding background, into an entangled representation with limited interpretability and fine-grained control~\cite{schutte2021using, cetin2023attri, wang2024disentangled}.

\subsection{Disentangled Representation Learning}

In contrast to holistic embedding-based approaches, various disentangled representation learning (DRL) studies have pursued a dimension-wise strategy for two main objectives. First, to derive more robust holistic embeddings with improved generalizability~\cite{dangovski2021equivariant, bai2023robust, devillers2023equimod, wang2024distortion, garrido2023self, song2024contrastive, wang2024disentangled}, and second, to explicitly decompose images into latent factors, each corresponding to distinct and semantically meaningful attributes, in order to improve explainability~\cite{burgess2018understanding, chen2016infogan, xiao2017dna, zhu2018visual}.

In medical imaging, the motivations for DRL are especially strong: improved interpretability, controllability, and robustness to confounding factors are crucial in clinical settings~\cite{liu2022learning, wang2022disentangled}. When transitioning DRL to medical imaging, researchers seek to decouple anatomical properties (\textit{e.g.}, organ size or shape)~\cite{wang2022semantic, jiang2022disentangled}, modality characteristics (\textit{e.g.}, CT and MRI sequences)~\cite{ouyang2021representation, chartsias2019disentangled}, and pathological factors (\textit{e.g.}, tumor presence or severity)~\cite{liu2022joint, hu2022clinically}, aiming to isolate distinct clinical features.

Despite notable progress, data- and perturbation-driven DRL approaches can be sensitive to training objectives, architecture choices, and assumptions about independent factors~\cite{locatello2019challenging, wang2024distortion, wang2024disentangled}. In existing studies, disentangled dimensions may not align precisely with real anatomical or pathological concepts, and spatial correspondence often remains weak when images are globally encoded. Consequently, there has been a growing emphasis on methods that combine disentanglement principles with controlled grouping or tokenization schemes to encapsulate semantically meaningful subsets of features.

\begin{figure*}[t]
\centering
\includegraphics[width=0.99\textwidth]{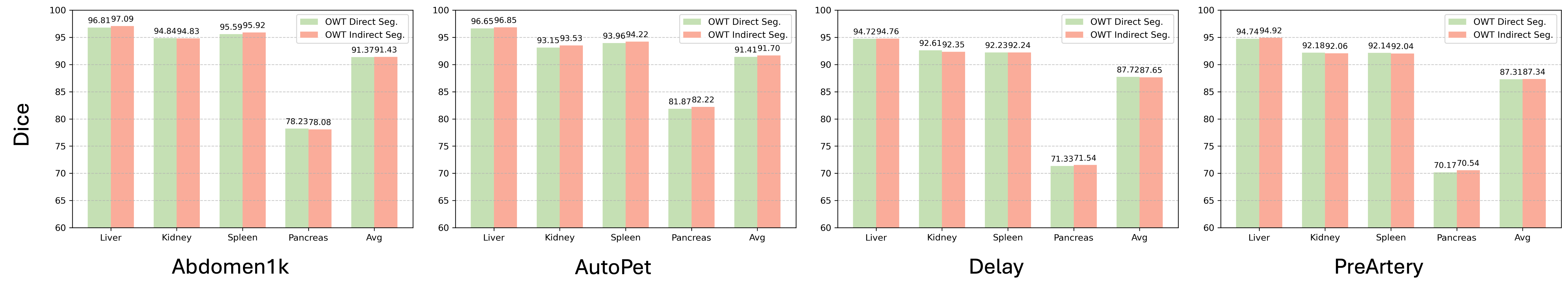}
\caption{Comparison of segmentation performance using direct and indirect approaches.} 
\label{fig: direct}
\end{figure*}

%%%%%%%%%%%%%%%%%%%%%%%%%%%%%%%%%%%%%%%%%%%%%%%%%%%%%%%%%%%%%%%%%%%%%%%%%%%%%%%%%%%%%
%%%%%%%%%%%%%%%%%%%%%%%%%%%%%%%%%%%%%%%%%%%%%%%%%%%%%%%%%%%%%%%%%%%%%%%%%%%%%%%%%%%%%
\section{Implementation Details}
\label{sec: imp}

\subsection{Image Preprocessing}
\label{subsec: pre}

For all 3D medical volumes, we standardize the spatial resolution to \(2.0 \text{mm} \times 2.0 \text{mm} \times 2.0 \text{mm}\) and apply a center crop of \(224 \times 224\). For CT datasets, we clip intensity values to the \([-1000, 1000]\) Hounsfield Unit (HU) range and normalize them to \([0,1]\). For MRI datasets, we clip intensities to the \([0, 99]\) percentile range and normalize them to \([0,1]\). We define five token groups to extract semantic features based on common annotations: background, liver, kidneys, spleen, and pancreas. Each medical image is further standardized in thickness by center-padding or cropping 112 slices based on the anatomical centers of the four organs. 

For the Abdomen1K dataset~\cite{ma2021abdomenct}, the number of cases in the training and test sets is 899:101. For AutoPet~\cite{Gatidis2022}, the split is 700:200, while for RAOS delay and pre-artery sequences~\cite{luo2024rethinking}, both have a training-to-test ratio of 330:83. All experiments, including OWT, MAE, and segmentation networks, are based on the same preprocessed dataset.

For the 2D face images in CelebAMaskHQ dataset~\cite{lee2020maskgan}, we resize them to \(224 \times 224\), and define six semantic entities based on annotations ({background}, {face\&ear}, {hair\&hat}, {mouth}, {nose}, and {eyes\&glasses}), as shown in Fig.~\ref{fig: facegen}. The CelebAMaskHQ dataset contains 30,000 facial image cases, with 24,000 images allocated for training and 6,000 for testing.

\subsection{Default Architecture of OWT and MAE}
\label{subsec: defarch}

In OWT, we adopt the same encoder and decoder architecture as MAE~\cite{he2022masked}, with the only modification being the replacement of the self-attention blocks~\cite{vaswani2017attention} with linear self-attention blocks~\cite{katharopoulos2020transformers}. The hyperparameters, optimizer and scheduler settings all follow MAE. By default, we use the ViT-base architecture, where the MAE encoder consists of 12 attention blocks, the decoder has 8, and the token embedding dimension is set to 768. For OWT, we configure the Encoder and Token Group Encoder with 6 attention blocks each, while keeping the decoder at 8 blocks. This ensures that the overall model length remains consistent with MAE. 

The default token number for each token group is 20 in OWT. Therefore, the training load of OWT is lower than that of MAE when using fewer than 10 SDTGs. To prevent information leakage during TGR training, we remove the class token from the OWT architecture. For OWT processing 2D images, we use a \(16 \times 16\) kernel for patch embedding. For 3D input volumes, we apply a \(1 \times 16 \times 16\) 3D kernel for patch embedding.

\begin{table}[t]
\centering
\small
\setlength{\tabcolsep}{1mm}
% \resizebox{0.99\columnwidth}{!}{
\begin{tabular}{lcccc}
\toprule
\multicolumn{1}{c}{\multirow{2}{*}{Method}} & \multicolumn{2}{c}{AutoPet} & \multicolumn{2}{c}{CelebAMaskHQ} \\
\cmidrule(lr){2-3} \cmidrule(lr){4-5}
\multicolumn{1}{c}{} & L2 $\downarrow$ & LPIPS $\downarrow$ & L2 $\downarrow$ & LPIPS $\downarrow$    \\
\midrule
% MAE~\cite{} & &               &             &              &           \\
MAE/$_\text{LPIPS}$ & 2.77e-04      & 0.0255      &  2.78e-03 & 0.1157      \\
OWT & 1.49e-04      & 0.0232      & 2.50e-03     & 0.1101    \\
OWT/$_\text{VQGAN}$ & \textbf{6.89e-05}      & \textbf{0.0073}      & \textbf{1.80e-03}     & \textbf{0.0540}   \\
\bottomrule
\end{tabular}
% }
\caption{Performance comparison when altering the Transformer-based encoder and decoder to those of VQGAN.}
\label{tab: gen}
\end{table}

\subsection{Experimental Setup}
\label{subsec: exset}

We follow the same experimental setup and hyperparameter settings as MAE. Additionally, each MAE- and OWT-based experiment is trained for 1200 epochs, with a 60-epoch warm-up phase. The training follows the learning rate protocol $\text{lr} = \text{base\_lr} \times {\text{effective\_batch\_size}} / {256}$ where the base learning rate is set to \(\text{base\_lr} = 10^{-4}\). 

For OWT with 2D image inputs, we set the batch size to 96. When processing 4-slice medical volumes, the batch size is set to 64. In medical imaging training, each unique input (whether a single slice or a 4-slice volume) is treated as an individual input. For instance, a 3D medical volume with a thickness of 112 slices is considered as 112 separate inputs during training. An epoch is defined as the point when all unique inputs in the dataset have been processed once.

For MAE training, we randomly mask 75\% of the tokens. During testing, we adjust the mask ratio to 0.25 for improved performance. When MAE is trained with LPIPS loss, all tokens are preserved for image reconstruction.

All reconstructed images are filtered with a 0.02 pixel-range threshold to reduce background noise. For the interpretability experiments validated by segmentation, we apply a 0.15 threshold to transfer the reconstructed results as semantic masks.

%%%%%%%%%%%%%%%%%%%%%%%%%%%%%%%%%%%%%%%%%%%%%%%%%%%%%%%%%%%%%%%%%%%%%%%%%%%%%%%%%%%%%
%%%%%%%%%%%%%%%%%%%%%%%%%%%%%%%%%%%%%%%%%%%%%%%%%%%%%%%%%%%%%%%%%%%%%%%%%%%%%%%%%%%%%

\section{Detailed Settings of Multi-Modality Incorporation}
\label{sec: multimodal}

% \subsection{OWT \textit{w/} Text Encoding}
% \subsection{Detailed Settings}
% \label{subsubsec: text}

For the OWT experiment incorporating text encoding, we design a text prompt inspired by \cite{liu2023clip}. Specifically, we define the template as \texttt{"a computerized tomography [category]"}, where \texttt{[category]} corresponds to \texttt{"background without the Liver, Kidneys, Spleen, and Pancreas"}, \texttt{"of a Liver"}, \texttt{"of Kidneys"}, \texttt{"of a Spleen"}, and \texttt{"of a Pancreas"}, aligning with the number of token groups. These 5 prompts are fed into a CT-specific pretrained CLIP text encoder, CT-CLIP~\cite{hamamci2024developing} to obtain the corresponding CT-consistent text encoding, resulting in a representation of size \(5 \times 512 \times 768\). Then, the mean is taken along the middle dimension, yielding a representation of size \(5 \times 768\).

To match the dimension token groups, we replicate each text encoding according to the number of tokens per token group (20), leading to a final text encoding dimension of \(100 \times 768\). Within the OWT framework, a linear transformation layer maps the text encoding from \(768\) to \(768\), resulting in a transformed text embedding of the same size \(100 \times 768\). This transformed text encoding is then element-wise added to the organ-wise representation \(X_G\) extracted by the Organ Collector. The combined representation proceeds through the random selection procedure in TGR, while all other components remain consistent with the default OWT settings.

% \subsection{Performance Analysis}
% In OWT, each organ is embedded into its corresponding SDTG, resulting in highly semantically disentangled and independent representations. Therefore, these representations can be efficiently aligned with or collaborated on with other modality embeddings (\textit{e.g.}, text encoding).

% Fig.~\ref{fig: text} and Table~\ref{tab: text} reveal that simply adding text prompts to organ-wise token groups instantly accelerates loss convergence and provides slight improvements in both holistic- and semantic-based reconstruction. Consequently, OWT shows promise for incorporating the text modality to further enhance organ-specific representations. Such multi-modal integration could unlock new medical applications, such as more robust organ labeling, guided retrieval and generation, and improved interpretability in complex workflows.

%%%%%%%%%%%%%%%%%%%%%%%%%%%%%%%%%%%%%%%%%%%%%%%%%%%%%%%%%%%%%%%%%%%%%%%%%%%%%%%%%%%%%
%%%%%%%%%%%%%%%%%%%%%%%%%%%%%%%%%%%%%%%%%%%%%%%%%%%%%%%%%%%%%%%%%%%%%%%%%%%%%%%%%%%%%

\section{Ablation Studies}
\label{sec: abl}

For the ablation study, we explore the following key aspects: the flexibility of the encoder-decoder architecture, the number of tokens per SDTG, and the impact of varying the amount of input data and labels in OWT.

\subsection{Backbone Alteration} 
\label{sec: BAlteration}
We investigate the effect of replacing the default ViT-based encoder-decoder architecture in OWT (adopted from MAE) with the encoder and decoder from VQGAN~\cite{esser2021taming}. Specifically, we configure the VQGAN bottleneck dimensions to \(14 \times 14\) for a \(224 \times 224\) input image, ensuring that the number of latent tokens (196) remains consistent with the ViT-based setting. As shown in Table~\ref{tab: gen}, despite the increased computational cost of VQGAN’s hybrid CNN and self-attention structure, it achieves significant performance gains over the transformer-based encoder-decoder. On two datasets, it improves both L2 and LPIPS evaluation metrics by 53.8\%–68.5\% in AutoPet and 43.8\%–51.0\% in CelebAMaskHQ, demonstrating the potential for further enhancements in OWT and highlighting its architectural flexibility.

\begin{figure}[t]
\centering
\includegraphics[width=0.49\textwidth]{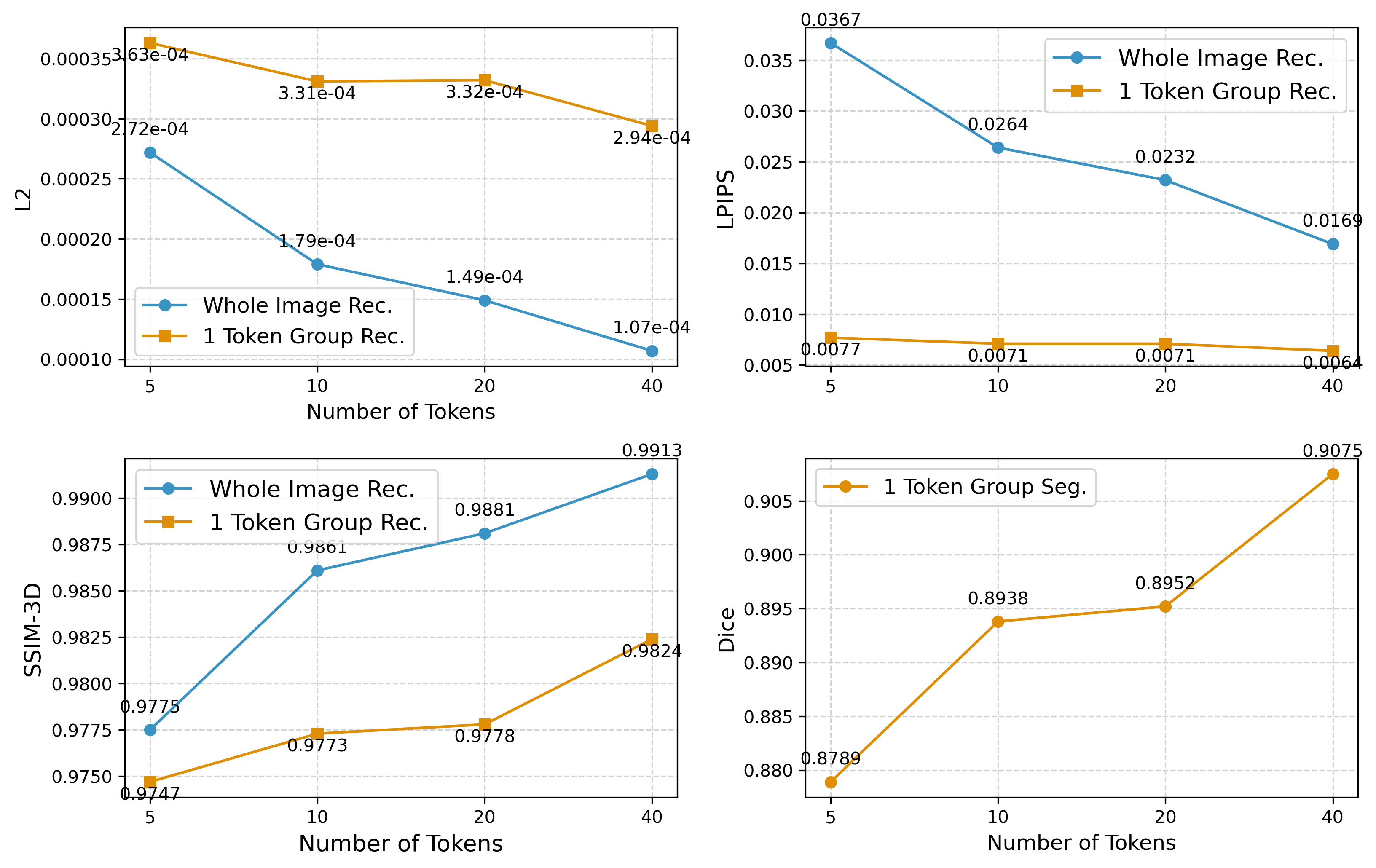}
\caption{Ablation study on the number of tokens in each token group when trained on AutoPet.} 
\label{fig: toknum}
\end{figure}

\begin{table}[t]
\centering
\small
\setlength{\tabcolsep}{1mm}
% \resizebox{0.7\columnwidth}{!}{
\begin{tabular}{lcccc}
\toprule
{AutoPet} & \multicolumn{2}{c}{Holistic Rec.} & \multicolumn{2}{c}{1 Token Group Rec.} \\
\cmidrule(lr){2-3} \cmidrule(lr){4-5}
OWT & L2 $\downarrow$ & LPIPS $\downarrow$ & L2 $\downarrow$ & LPIPS $\downarrow$ \\
\midrule
Regular & 1.49e-04 & 0.0232 & 3.32e-04 & 0.0071 \\
Adaptive & \textbf{1.40e-04}  & \textbf{0.0215} & \textbf{3.19e-04} & \textbf{0.0067} \\
\bottomrule
\end{tabular}
% }
\caption{Reconstruction comparison of OWT with regular and adaptively distributed SDTGs.}
\label{tab: adaptivedis}
\end{table}

\subsection{Number of Tokens per SDTG} 
The number of tokens within each SDTG determines its capacity to store organ-specific information. To evaluate this, we experiment with SDTG token counts of 5, 10, 20, and 40, assessing their impact on both generation and segmentation tasks. For the generation task, we measure full-image reconstruction using all SDTGs and single-organ reconstruction using only one SDTG, evaluated by L2, LPIPS, and SSIM-3D metrics. For segmentation, we assess the Dice score of single-organ masks.

As shown in Fig.~\ref{fig: toknum}, increasing the token count from 10 to 20 provides a slight performance improvement, while further increasing it from 20 to 40 leads to a more significant boost. This suggests that OWT has considerable scalability and potential as a foundational semantic-level tokenization framework, capable of further enhancements with increased token capacity.

\begin{figure}[t]
\centering
\includegraphics[width=0.49\textwidth]{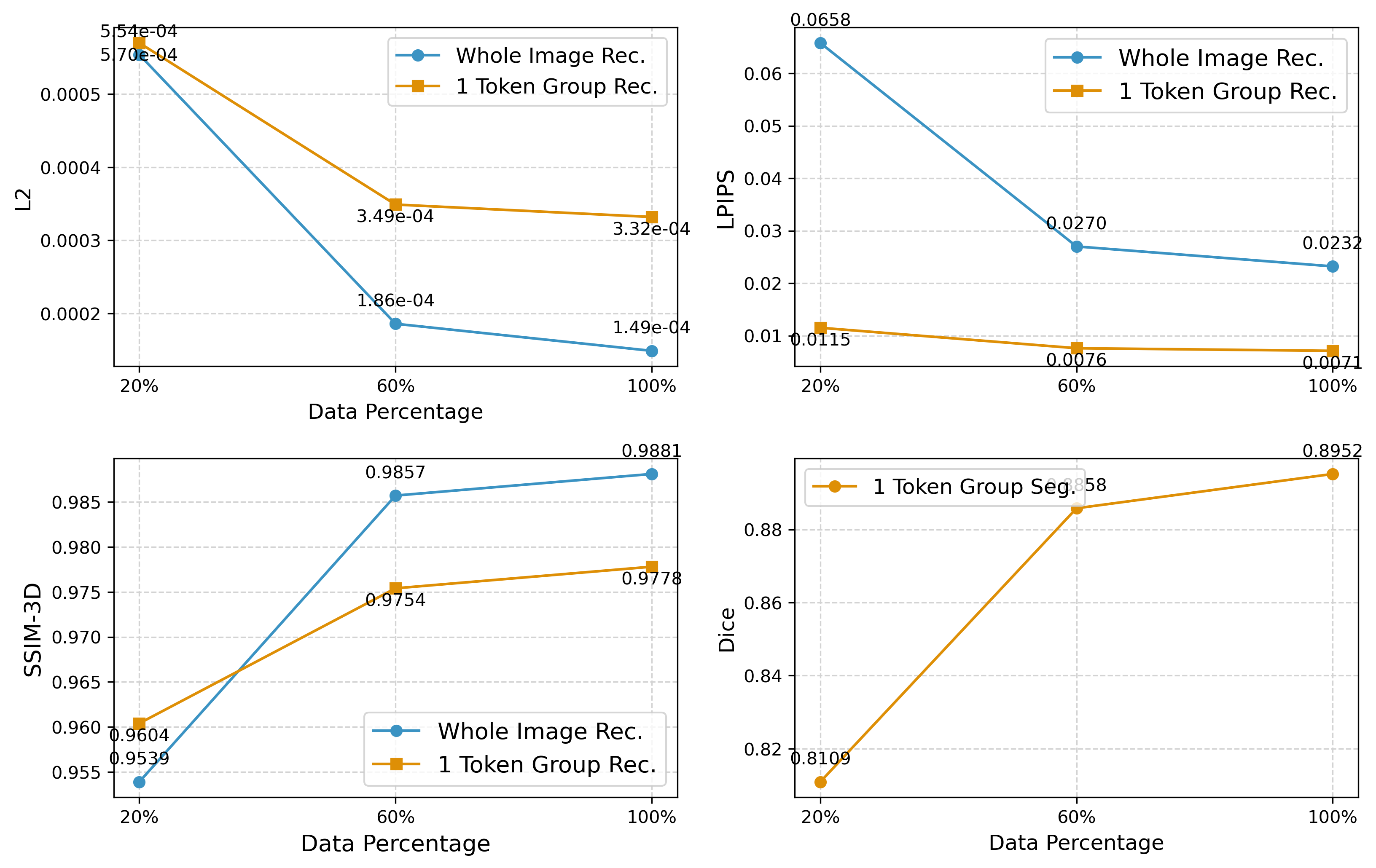}
\caption{Ablation study on OWT performance when utilizing different percentages of the dataset (AutoPet).} 
\label{fig: percedata}
\end{figure}

\begin{table*}[!h]
\begin{minipage}[b]{.48\textwidth}
  \centering
  \small
\setlength{\tabcolsep}{1mm}
% \resizebox{0.8\columnwidth}{!}{
\begin{tabular}{lccccl}
\toprule
Abdomen1k   & \multicolumn{1}{c}{Liver} & \multicolumn{1}{c}{Kidney} & \multicolumn{1}{c}{Spleen} & \multicolumn{1}{c}{Pancreas} & \multicolumn{1}{l}{Avg Dice $\uparrow$} \\
\midrule
% & Dice   & Dice   & Dice   & Dice & Dice  \\
UNetplus       & 96.85          & 94.63          & 91.00          & 71.05          & 88.38          \\
AttU\_Net      & \textbf{97.09} & \textbf{95.22} & 93.05          & 72.00          & 89.34          \\
TransUnet      & 96.78          & 94.65          & 92.30          & 71.89          & 88.91          \\
SwinUnet       & 96.53          & 93.68          & 90.52          & 65.50          & 86.56          \\
CMUNet         & 96.05          & 94.67          & 90.84          & 71.75          & 88.33          \\
\midrule
OWT & 96.55          & 94.21          & 95.12          & 71.32          & 89.30          \\
OWT/$_{\text{4-slice}}$ & 96.81          & 94.84          & \textbf{95.59} & \textbf{78.23} & \textbf{91.37}{$_{+2.03}$} \\
\toprule
\toprule
AutoPet   & \multicolumn{1}{c}{Liver} & \multicolumn{1}{c}{Kidney} & \multicolumn{1}{c}{Spleen} & \multicolumn{1}{c}{Pancreas} & \multicolumn{1}{l}{Avg Dice $\uparrow$} \\
\midrule
% & Dice   & Dice   & Dice   & Dice & Dice  \\
UNetplus       & 96.41          & 92.56          & 91.31          & 76.87          & 89.29          \\
AttU\_Net      & 96.60          & 92.86          & 91.22          & 78.24          & 89.73          \\
TransUnet      & 96.34          & \textbf{93.59} & 90.65          & 78.23          & 89.70          \\
SwinUnet       & 96.33          & 92.34          & 91.11          & 72.46          & 88.06          \\
CMUNet         & 96.56          & 93.11          & 90.61          & 77.65          & 89.48          \\
\midrule
OWT & 96.14          & 92.33          & 92.79          & 76.82          & 89.52          \\
OWT/$_{\text{4-slice}}$ & \textbf{96.65} & 93.15          & \textbf{93.96} & \textbf{81.87} & \textbf{91.41}{$_{+1.68}$} \\
\bottomrule
\end{tabular}
% }
\captionof{table}{Performance of semantic segmentation on CT datasets.}
\label{tab: seg_ct}
\end{minipage}\quad
\begin{minipage}[b]{.48\textwidth}
  \centering
  \small
\setlength{\tabcolsep}{1mm}
% \resizebox{0.8\columnwidth}{!}{
\begin{tabular}{lccccl}
\toprule
Delay   & \multicolumn{1}{c}{Liver} & \multicolumn{1}{c}{Kidney} & \multicolumn{1}{c}{Spleen} & \multicolumn{1}{c}{Pancreas} & \multicolumn{1}{l}{Avg Dice $\uparrow$} \\
\midrule
% & Dice   & Dice   & Dice   & Dice & Dice  \\
UNetplus       & 93.53          & 91.85          & 85.20          & 65.85          & 84.11          \\
AttU\_Net      & 94.25          & 91.08          & 86.85          & 65.98          & 84.54          \\
TransUnet      & 93.93          & 90.83          & 89.07          & 66.03          & 84.97          \\
SwinUnet       & 92.78          & 90.61          & 86.35          & 62.55          & 83.07          \\
CMUNet         & 94.02          & 91.88          & 87.74          & 65.64          & 84.82          \\
\midrule
OWT & 94.32          & 91.87          & 91.55          & 65.65          & 85.85          \\
OWT/$_{\text{4-slice}}$ & \textbf{94.72} & \textbf{92.61} & \textbf{92.23} & \textbf{71.33} & \textbf{87.72}{$_{+1.87}$} \\
\toprule
\toprule
PreArtery   & \multicolumn{1}{c}{Liver} & \multicolumn{1}{c}{Kidney} & \multicolumn{1}{c}{Spleen} & \multicolumn{1}{c}{Pancreas} & \multicolumn{1}{l}{Avg Dice $\uparrow$} \\
\midrule
% & Dice   & Dice   & Dice   & Dice & Dice  \\
UNetplus       & 94.05          & 92.25          & 87.26          & 64.14          & 84.42          \\
AttU\_Net      & 94.33          & 92.32          & 89.07          & 64.45          & 85.04          \\
TransUnet      & 94.33          & 90.69          & 86.59          & 65.56          & 84.29          \\
SwinUnet       & 92.91          & 88.91          & 86.33          & 61.48          & 82.41          \\
CMUNet         & 94.26          & \textbf{92.58} & 89.21          & 66.18          & 85.56          \\
\midrule
OWT & 94.19          & 91.10          & 91.82          & 65.57          & 85.67          \\
OWT/$_{\text{4-slice}}$ & \textbf{94.74} & 92.18          & \textbf{92.14} & \textbf{70.17} & \textbf{87.31}{$_{+1.64}$} \\
\bottomrule
\end{tabular}
% }
\captionof{table}{Performance of semantic segmentation on MRI datasets.}
\label{tab: seg_mr}
\end{minipage}
\end{table*}

\subsection{Adaptively Distributed SDTGs} 
\label{sec: distribute}

To explore this, we include a proof-of-concept experiment showing that the OWT framework has the potential to support adaptive token group sizing. We allocated token numbers to different organs based on the fourth root of their volumes. For background (containing all information other than defined organs), liver, kidney, spleen, and pancreas, 45, 20, 13, 13, and 9 were allocated, with the total tokens fixed at 100 (identical to the default setting). Preliminary results show that this adaptive allocation better accommodates the varying feature distributions across organs and provides performance improvements over a fixed-size setting (Table~\ref{tab: adaptivedis}), demonstrating the scalability of OWT as a novel foundational tokenization framework.

\subsection{Data Percentage} 
Fig.~\ref{fig: percedata} illustrates the performance of the OWT framework on the AutoPet dataset when trained with 20\%, 60\%, and 100\% of the data. We observe a significant improvement in performance when increasing the dataset size from 20\% to 60\%, whereas the gain from 60\% to 100\% is relatively marginal. This suggests that OWT efficiently learns meaningful representations with a moderate amount of data, indicating its strong data efficiency and the potential for effective generalization even with limited training samples.

%%%%%%%%%%%%%%%%%%%%%%%%%%%%%%%%%%%%%%%%%%%%%%%%%%%%%%%%%%%%%%%%%%%%%%%%%%%%%%%%%%%%%
%%%%%%%%%%%%%%%%%%%%%%%%%%%%%%%%%%%%%%%%%%%%%%%%%%%%%%%%%%%%%%%%%%%%%%%%%%%%%%%%%%%%%

\begin{figure}[t]
\centering
\includegraphics[width=0.50\textwidth]{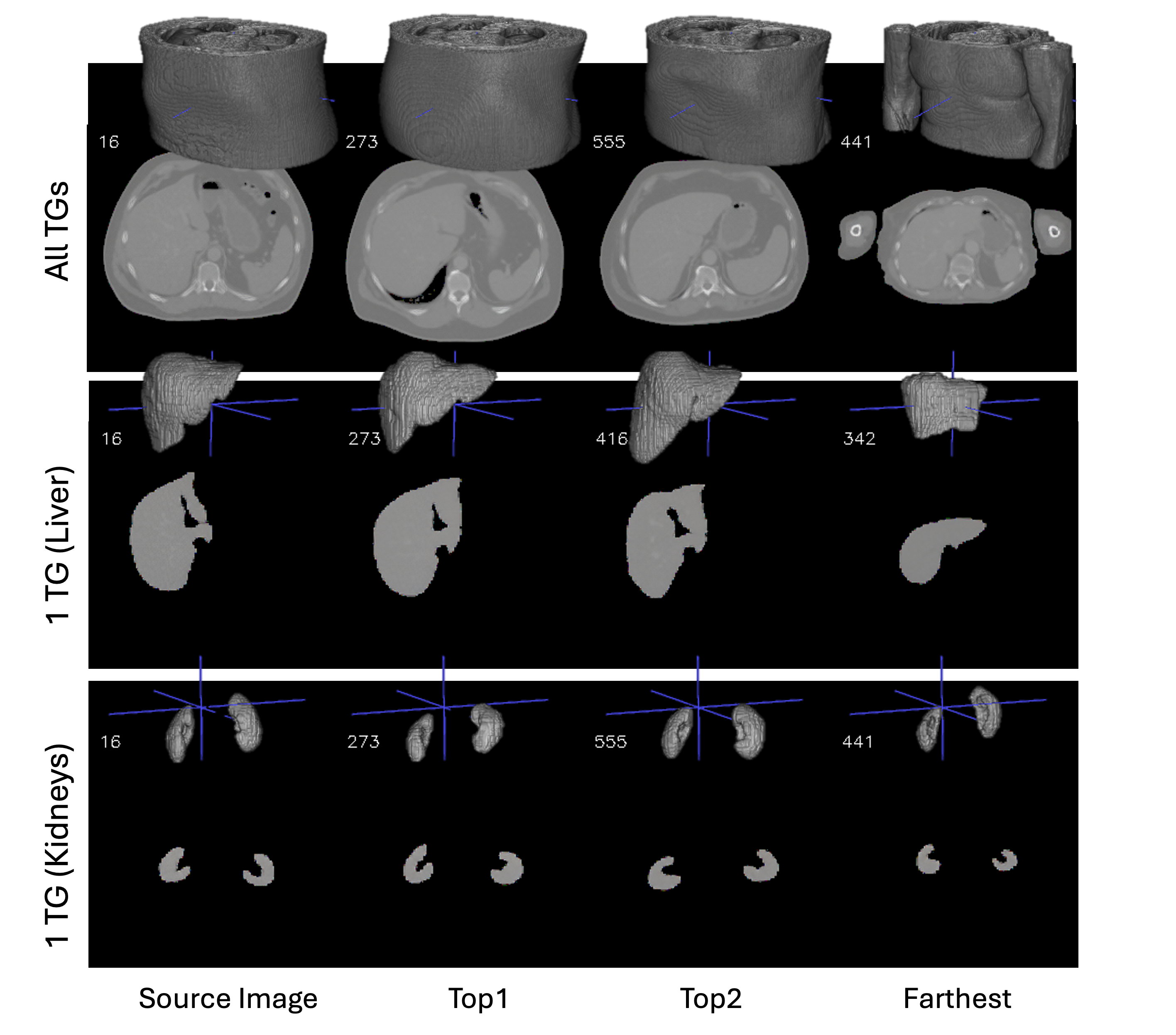}
\caption{Semantic-based retrieval of CT images.} 
\label{fig: medretct}
\end{figure}

\section{Detailed Semantic Segmentation Results}
\label{sec: seg}

\noindent
\textbf{Detailed Results.}
Tables~\ref{tab: seg_ct} and~\ref{tab: seg_mr} provide a more detailed results of those presented in Table~\ref{tab: seg_ct1}, showing the Dice scores for each of the four organs across different datasets. From Tables~\ref{tab: seg_ct} and~\ref{tab: seg_mr}, we observe that compared to dedicated segmentation networks, OWT/$_{\text{4-slice}}$ generally achieves superior performance across all organs. Notably, for smaller organs such as the pancreas, OWT/$_{\text{4-slice}}$ demonstrates a significant improvement, with an advantage exceeding a 5.3 dice score in the best case.

\noindent
\textbf{Token Group Connectivity.}
For semantic segmentation using OWT, we generate masks by processing independently utilized Semantically Disentangled Token Groups (SDTGs) through the Token Group Encoder, AHER, and decoder to reconstruct an image, which is then thresholded to produce a segmentation mask. We refer to this mask as the ``direct" mask. Since SDTGs can be flexibly combined, we also explore an alternative approach: subtracting the reconstructed images from all other four SDTGs from the input image and applying a threshold to obtain a predicted mask for the same organ. We define this as the ``indirect" mask.

Although using all complementary SDTGs reduces computational efficiency compared to single SDTG usage, it incorporates additional inter- and intra-organ information, potentially enhancing connectivity in the segmentation process. To evaluate whether leveraging more SDTGs improves boundary and shape reconstruction, we conduct comparative experiments. The results, presented in Fig.~\ref{fig: direct}, show that across four datasets, the direct and indirect methods yield largely similar outcomes. This demonstrates that the SDTGs extracted by OWT maintain a high degree of semantic independence, as they accurately capture organ boundaries and shapes even when used independently. These findings align with the t-SNE visualization of SDTG relationships in Fig.~\ref{fig: tsne4}, further validating their disentangled nature.

\section{Additional Organ-Level Retrieval Demonstration}
\label{sec: retrict}

An additional example demonstrates OWT performing organ-level retrieval on the CT modality (Fig.~\ref{fig: medretct}). Combined with Fig.~\ref{fig: medret}, this shows the stability and potential of OWT for organ-level retrieval across different modalities. It can not only retrieve patients with similar body shapes at a holistic level, but also retrieve organs in similar conditions and locations for specific organs of interest. This highlights the more flexible and unique clinical value of OWT, which cannot be achieved by existing retrieval methods based on holistic embeddings.

\end{document}